%% file: main.tex
\pdfoutput=1

\documentclass[11pt]{article}
\usepackage[final]{acl}
\usepackage{times}
\usepackage{latexsym}
\usepackage[T1]{fontenc}
\usepackage[utf8]{inputenc}
\usepackage{microtype}
\usepackage{inconsolata}

\usepackage{xspace}
\usepackage{multirow}
\usepackage{booktabs}
\usepackage{bm}
\usepackage{amsmath}
\usepackage{amsfonts}
\usepackage{hyperref}
\usepackage{graphicx}
\usepackage{framed}
\usepackage{enumitem}
\usepackage{pifont}
\usepackage{CJKutf8}
\usepackage{marvosym}
\usepackage{lipsum}
\usepackage{CJKutf8}
\usepackage{pmboxdraw}
\usepackage{caption}
\usepackage{subcaption}
\usepackage{epigraph}
\usepackage{longtable}

\newcommand{\ie}{\emph{i.e.,}\xspace}

\newcommand{\eg}{\emph{e.g.,}\xspace}

\newcommand{\etc}{\emph{etc}}
\newcommand{\ignore}[1]{}

\makeatletter
\def\@fnsymbol#1{}
\makeatother

\title{Language-Specific Neurons: \\ The Key to Multilingual Capabilities in Large Language Models}

\author{
Tianyi Tang\textsuperscript{\rm{1 * \dag}}\thanks{*\ This work was done during internship at MSRA.}, 
Wenyang Luo\textsuperscript{\rm{1 \dag}}\thanks{\dag\ Equal contribution.}, 
Haoyang Huang\textsuperscript{\rm{2}}, 
Dongdong Zhang\textsuperscript{\rm{2}} \\ 
\textbf{Xiaolei Wang}\textsuperscript{\rm{1}}\textbf{,}
\textbf{Wayne Xin Zhao}\textsuperscript{\rm{1 \Letter}\thanks{\textsuperscript{\Letter}\ Corresponding author}\ }\textbf{,}
\textbf{Furu Wei}\textsuperscript{\rm{2}}\textbf{,}
\textbf{Ji-Rong Wen}\textsuperscript{\rm{1,3}} \\
\textsuperscript{1} Gaoling School of Artificial Intelligence, Renmin University of China \\
\textsuperscript{2} Microsoft Research Asia, China \\
\textsuperscript{3} School of Information, Renmin University of China \\
\texttt{\{steventianyitang,wengyang\_luo\}@outlook.com \ \ wxl1999@foxmail.com} \\
\texttt{\{haohua,dozhang,fuwei\}@microsoft.com \ \ batmanfly@gmail.com} \\
}

\begin{document}
\maketitle
\begin{abstract}
Large language models (LLMs) demonstrate remarkable multilingual capabilities without being pre-trained on specially curated multilingual parallel corpora.
It remains a challenging problem to explain the underlying mechanisms by which LLMs process multilingual texts.
In this paper, we delve into the composition of Transformer architectures in LLMs to pinpoint language-specific regions.
Specially, we propose a novel detection method, language activation probability entropy~(\emph{LAPE}), to identify language-specific neurons within LLMs.
Based on LAPE, we conduct comprehensive experiments on several representative LLMs, such as LLaMA-2, BLOOM, and Mistral. Our findings indicate that LLMs' proficiency in processing a particular language is predominantly due to a small subset of neurons, primarily situated in the models' top and bottom layers.
Furthermore, we showcase the feasibility to ``steer'' the output language of LLMs by selectively activating or deactivating language-specific neurons. 
Our research provides important evidence to the understanding and exploration of the multilingual capabilities of LLMs. 

\end{abstract}

\input{sec-intro}
\input{sec-method}

\input{sec-exp}

\input{sec-analysis}

\input{sec-rel}

\section{Conclusion}
Despite the impressive multilingual capabilities demonstrated by LLMs, the understanding of how these abilities develop and function remains nascent. In this paper, we introduced a novel detection method, \ie language activation probability entropy~(\emph{LAPE}), to pinpoint language-specific neurons within LLMs. LAPE assesses the response of individual neurons to various languages, selecting those with a propensity for activation when exposed to one or two languages. Based on LAPE, we further conducted extensive experiments to investigate the multilingual capabilities of LLMs. Specially, we have found that an LLM’s proficiency in processing different languages is significantly influenced by a small subset of neurons, which are mainly located in the model's top and bottom layers. We have further demonstrated that the output language of LLMs can be directed by selectively enabling or disabling these language-specific neurons.
For future work, we aim to leverage these findings to enhance knowledge transfer between major and minor languages and  devise efficient strategies for continual pre-training to better accommodate specific languages.

\section*{Acknowledgement}
This work was partially supported by National Natural Science Foundation of China under Grant No. 62222215, Beijing Natural Science Foundation under Grant No. L233008 and 4222027. Xin Zhao is the corresponding author.

\section*{Limitations}
In this study, we employ language activation probability entropy as a metric to identify language-specific neurons. However, it is important to note that our method is relative to the presence of multiple languages. In scenarios where only a single language is present, establishing an absolute threshold to determine the language-relatedness of neurons is not feasible.
Moreover, the criteria for distinguishing between high-resource and low-resource languages within the model warrant further investigation. The model's possibility to managing a large number of languages, as well as the differences between various languages, represents a promising avenue for future research.
Finally, our research has only begun to explore the possibility for directing the output language of the model. Developing strategies to harness these identified neurons for enhancing the model's multilingual proficiency is still worth exploring.
\bibliography{ref}
\input{sec-app}

\end{document}

%% file: sec-intro.tex
\section{Introduction}

\epigraph{\emph{The brain has its own language for testing the structure and consistency of the world.}}{Carl Sagan
}




The pursuit of multilingual capabilities, mirroring our world's linguistic diversity, is a critical research objective that paves the way for information democratization  across linguistic divides. 
The emergence of pre-trained language models (PLMs) such as mBERT~\cite{devlin-etal-2019-bert} and XLM-R~\cite{conneau-etal-2020-unsupervised} has marked a significant shift towards enhanced multilingual understanding.
Furthermore, large language models (LLMs), such as GPT-4~\cite{achiam-etal-2023-gpt} and PaLM-2~\cite{anil-etal-2023-palm}, 
have recently demonstrated more excellent  multilingual capabilities in language understanding, reasoning, and generation, despite being predominantly trained in English corpora.


\begin{figure}[!t]
\centering
\includegraphics[width=1.0\columnwidth]{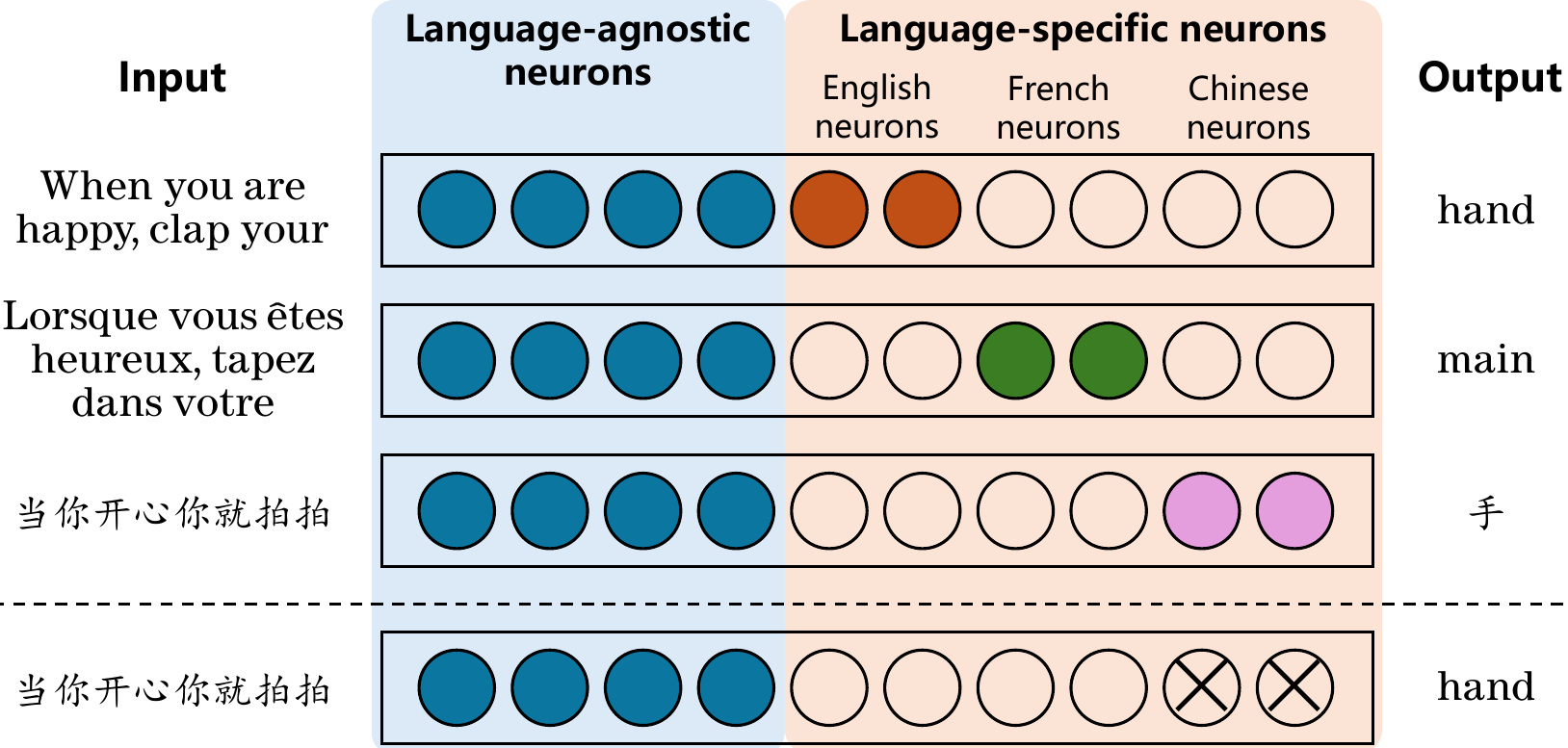}
\caption{An illustration of region distribution of activated neurons when predicting the next word in language models across different languages. Here, colored circles denote activated neurons. When Chinese-specific neurons are deactivated (denoted by $\otimes$), the model may produce outputs in English. }
\label{fig:model}
\end{figure}


Existing studies~\cite{pires-etal-2019-multilingual,conneau-etal-2020-emerging} have mainly explored how multilingual PLMs (\eg mBERT) possess semantic alignment capabilities across languages despite the absence of multilingual parallel corpora.
They have identified several critical factors that influence cross-lingual transfer, including training data (\eg overlapped tokens) and training settings (\eg shared parameters)~\cite{dufter-schutze-2020-identifying,philippy-etal-2023-towards}.
Nevertheless, the underlying mechanisms by which the model itself process diverse languages at the composition level continue to be an area of vigorous investigation.  

To develop a deeper understanding of the multilingual capabilities of LLMs, we draw inspiration from the neurobiological underpinnings of human language faculties~\cite{friederici-etal-2011-brain,parr-etal-2022-active,khanna-etal-2024-single}. 
Specific regions within the human brain, such as Broca's area and Wernicke's area have been identified to support particular language functions.
To make an analogy with human's language functions, we posit that regions within the language models can be delineated into two primary components: \emph{language-agnostic regions} that encompass universal knowledge and pragmatics principles, and \emph{language-specific regions} that handle language-specific vocabulary, grammar, and idiomatic expressions.
Figure~\ref{fig:model} presents such a conceptual illustration of region distribution in LLMs posited by us. Actually, language-agnostic regions have been widely explored in existing literature, including knowledge storing~\cite{dai-etal-2022-knowledge} and task handling~\cite{wang-etal-2022-finding-skill}. However,  language-specific regions, especially those supporting multilingual capacities, have been seldom studied in existing literature of of LLMs, which is the focus of our research. 

In this work, we first propose a novel detection method called \emph{language activation probability entropy~(LAPE)} to identify \textbf{language-specific neurons} within LLMs. This method involves computing the activation likelihood of individual neurons in response to corpora across different languages. Subsequently, we select neurons with lower language activation probability entropy as language-specific neurons, \ie those having a higher activation probability for one or two particular languages and a lower probability for others. 
Furthermore, based on the proposed LAPE method, we have conducted a systematic study with language-specific regions of two popular open-sourced LLMs, leading to several major findings: 

$\bullet$ First, the proficiency of an LLM in processing a particular language can be significantly impacted by a \emph{minuscule proportion} of its neurons. Deactivating such language-specific neurons leads to a remarkable degradation in the model's understanding and generation abilities for that language.


$\bullet$ Second, neurons specific to individual languages are predominantly located in the \emph{bottom} and \emph{top} layers of LLMs. The bottom layers mainly serve to process the input from various languages into the unified semantic space of a high-resource language (\eg English), while the top layers  project the semantic content (after the processing of middle layers) into the respective tokens in the corresponding vocabulary of each language. 

$\bullet$ Third, we demonstrate the potential to ``steer'' the output language of LLMs by selectively activating and/or deactivating certain neurons. Our approach could provide a promising solution to mitigate the off-target issue (\eg the tendency of LLaMA-2 to reply in English to Chinese queries), while stimulating the capabilities of cross-lingual generation tasks.

To the best of our knowledge, it is the first study that investigates language-specific regions inside LLMs and analyzes the how these regions influence LLMs' capabilities to process multilingual texts. We introduce the concept of ``language-specific neurons'' and propose language activation probability entropy to identify such neurons in LLMs. 
We make available the identified language-specific neurons and corresponding code at \url{https://github.com/RUCAIBox/Language-Specific-Neurons}.

%% file: sec-method.tex
\section{Identifying Language-Specific Regions}

\subsection{Background}
Currently, LLMs are predominantly developed on an auto-regressive Transformer architecture~\cite{Vaswani-etal-2017-attention}, in which the basic building blocks are the multi-head self-attention (MHA) and the feed-forward network (FFN).
Given the hidden state $\bm{h}^{i-1} \in \mathbb{R}^{d}$ of ($i-1$)-th layer of a specific token, the MHA module inside the $i$-th layer can be expressed as follows:
\begin{equation} \label{eq-1}
    \tilde{\bm{h}}^i = \text{Attn}(\bm{h}^{i-1}\bm{W}^i_q, \bm{H}^{i-1}\bm{W}^i_k, \bm{H}^{i-1}\bm{W}^i_v) \cdot \bm{W}^i_o,
\end{equation}
where $\bm{W}^i_q$, $\bm{W}^i_k$, $\bm{W}^i_v$, and $\bm{W}^i_o$ represent the trainable parameters, and $\bm{H}^{i-1}$ stands for the hidden states in the previous layer of the whole sequence. Subsequently, the FFN module is described by the following formulation:
\begin{equation} \label{eq-2}
    \bm{h}^i = \text{act\_fn}(\tilde{\bm{h}}^i\bm{W}^i_1) \cdot \bm{W}^i_2,
\end{equation}
where $\bm{W}^i_1 \in \mathbb{R}^{d \times 4d}$ and $\bm{W}^i_2 \in \mathbb{R}^{4d \times d}$ are parameters and $\text{act\_fn} (\cdot)$ denotes the activation function (\eg  GELU~\cite{hendrycks-etal-2016-gaussian} for BLOOM~\cite{workshop-etal-2022-bloom}). 
A \emph{neuron} is defined as a linear transformation of a single column in $\bm{W}^i_1$ followed by a non-linear activation.
Consequently, a FFN module within a single layer consists of $4d$ neurons.
As a new variant of activation function,  GLU~\cite{shazeer-etal-2020-glu} has been widely used in recent LLMs (\eg LLaMA~\cite{touvron-etal-2023-llama}) for  improving the performance of Transformer:
\begin{equation} \label{eq-3}
    \bm{h}^i = (\text{act\_fn}(\tilde{\bm{h}}^i\bm{W}^i_1) \otimes \tilde{\bm{h}}^i\bm{W}^i_3)\cdot \bm{W}^i_2.
\end{equation}
In our work, the $j$-th neuron inside the $i$-th FFN layer is considered to be \emph{activated} if its respective activation values $\text{act\_fn}(\tilde{\bm{h}}^i\bm{W}^i_1)_j$ exceed zero~\cite{Nair-etal-2010-rectified}.


\subsection{Language Activation Probability Entropy} \label{subsec-method}
In existing research, neurons within the FFN modules are found to be capable of storing factual knowledge~\cite{dai-etal-2022-knowledge}, encoding positional information~\cite{voita-etal-2023-neurons}, responding to particular syntactic triggers~\cite{gurnee-etal-2024-universal}, \etc. Inspired by these findings, we posit that there exist specific neurons in LLMs for multilingual processing. Next, we introduce a new detection method based on language activation probability entropy (\emph{LAPE}) to identify language-specific  neurons.

Our research primarily focuses on pre-trained foundation models (\eg LLaMA-2 and BLOOM), rather than fine-tuned models that have undergone instruction tuning or RLHF, which helps reduce other influencing factors.
Specially, we feed existing LLMs with multilingual texts, each written in a single language. 
For the $j$-th neuron in the $i$-th layer, we then compute the \emph{activation probability} when processing texts in language $k$:
\begin{equation} \label{eq-4} 
p^k_{i,j} = \mathbb{E}\left(\mathbb{I}(\text{act\_fn}(\tilde{\bm{h}}^i \bm{W}^i_1)_j > 0) \mid \text{language } k \right),
\end{equation}
where $\mathbb{I}$ is the indicator function.
The activation probability is empirically estimated by the likelihood that the neuron's activation value exceeds zero.
Subsequently, we can obtain the distribution $\bm{p}_{i,j} = (p^1_{i,j}, \dots, p^k_{i,j}, \dots, p^l_{i,j})$ for each neuron, indicating its probability of activation for each language.
To convert $\bm{p}_{i,j}$ into a valid probability distribution, we apply L1 normalization, yielding $\bm{p}_{i,j}'$. The entropy of this distribution, which we refer to as \emph{language activation probability entropy}, is computed to quantify the neuron's language activation reaction:
\begin{equation} \label{eq-5}
    \text{LAPE}_{i,j} = -\sum_{k=1}^l p'^{k}_{i,j} \log (p'^{k}_{i,j}).
\end{equation}
We designate neurons with low LAPE scores as ``\textbf{language-specific neurons}'', as they demonstrate a predilection for activation in response to one or two languages, while showing reduced activation probabilities for others.

In implementation, we collect multilingual corpora sourced from Wikipedia, a widely recognized and high-quality resource for diverse languages, and sample documents to create a dataset comprising 100 million tokens for each language. Subsequently, we input these tokens into a target LLM and follow Equations~\ref{eq-4} and~\ref{eq-5} to compute the LAPE score for individual neurons. Finally, we select neurons that fall within the lowest percentile of LAPE scores, specifically targeting the bottom 1\%. To refine our selection, we further impose a predefined threshold to exclude neurons exhibiting negligible activation probability:  a neuron is deemed specific to language $k$ if its corresponding activation probability $p^{k}_{i,j}$ surpasses the threshold.










%% file: sec-exp.tex
\section{Experiments}

In this section, we present empirical evaluation to substantiate the efficacy of our proposed LAPE method and elucidate the impact of language-specific neurons on  multilingual capacities.

\subsection{Experimental Setup}\label{sec:exp-setup}
\paragraph{Models.}
We conducted our study primarily on two publicly available large language models (LLMs): LLaMA-2~\cite{touvron-etal-2023-llama2} and BLOOM~\cite{workshop-etal-2022-bloom}. Among them, LLaMA-2 is recognized for its excellence as a foundational model, primarily pre-trained on English texts, while BLOOM is noted for its multilingual proficiency due to a balanced distribution of training languages. Specifically, we investigate multiple versions of LLaMA-2: the 7B, 13B, and 70B models, which contain approximately 352K, 553K, and 2.29M neurons, respectively. For BLOOM, we select the 7.1B version, consisting of roughly 492K neurons. The languages we focus on include English (\emph{en}), Simplified Chinese (\emph{zh}), French (\emph{fr}), Spanish (\emph{es}), Vietnamese (\emph{vi}), Indonesian (\emph{id}), and Japanese (\emph{ja}). We exclude Japanese (\emph{ja}) for BLOOM since it has not been pre-trained on Japanese corpora. 
To verify the generality of our method, we also include LLMs under different settings, including LLaMA-2 Chat, OPT~\cite{zhang-etal-2022-opt}, Mistral~\cite{jiang-etal-2023-mistral}, and Phi-2~\cite{javaheripi-etal-2023-phi}.

\paragraph{Dataset.}
Our analysis of language-specific neurons is conducted across two distinct dimensions:

$\bullet$ \emph{Language modeling}: We assess the multilingual language modeling capability using perplexity (PPL) scores on Wikipedia corpora. Our dataset comprises one million tokens per language, all sourced after September 2022 to ensure the content has not been included in the training sets of either LLaMA-2 or BLOOM. 

$\bullet$ \emph{Open-ended generation}: 
To evaluate the model's multilingual generation capabilities in real-world scenarios, we translate a set of questions from the Vicuna dataset~\cite{weilin-etal-2023-vicuna} into target languages using \texttt{gpt-4-0125-preview}. The questions span a broad spectrum of topics, deliberately excluding mathematics and coding queries to maintain a focus on language processing proficiency and avoid confounding variables. We utilize greedy search with a repetition penalty of 1.1 to generate output. The resulting texts are then assessed by GPT-4 on a scale ranging from 1 to 10, following the methodology described by~\citet{zheng-etal-2023-judging}.

\begin{figure}[t!]
\centering
\begin{subfigure}{0.49\linewidth}
\includegraphics[width=\linewidth]{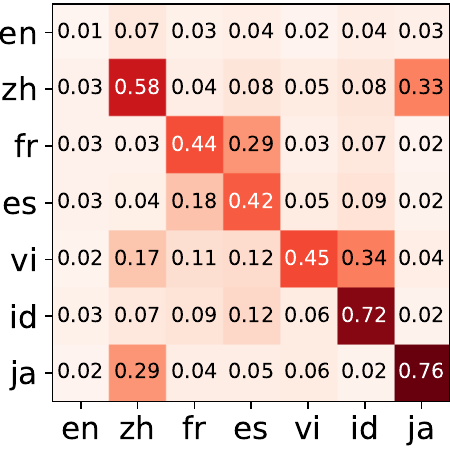}
\caption{LAPE}
\end{subfigure}
\begin{subfigure}{0.49\linewidth}
\includegraphics[width=\linewidth]{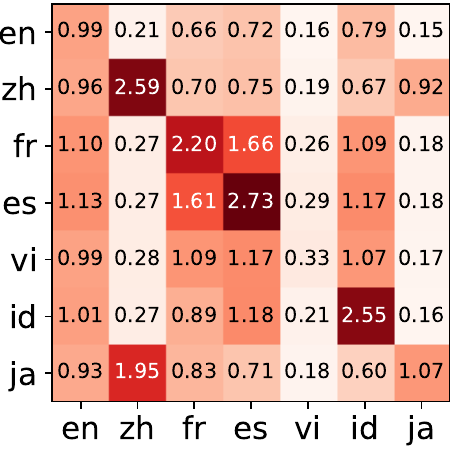}
\caption{LAP}
\end{subfigure}
\begin{subfigure}{0.49\linewidth}
\includegraphics[width=\linewidth]{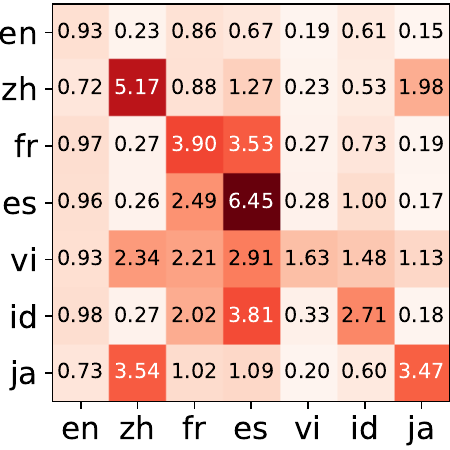}
\caption{LAVE}
\end{subfigure}
\begin{subfigure}{0.49\linewidth}
\includegraphics[width=\linewidth]{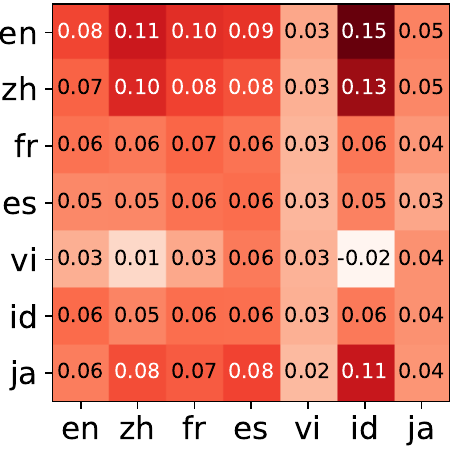}
\caption{PV}
\end{subfigure}
\caption{Impact of four identification methods on the PPL increase of LLaMA-2 (7B). The element at the $i$-th row and $j$-th column is the PPL change for language $j$ due to perturbations in a specific region of language~$i$.}
\label{fig:ppl-results1}
\end{figure}

\paragraph{Identification methods.} 
For comparison, we consider the following methods for identifying language-specific regions: 

(a) \emph{Language activation probability entropy (LAPE, ours)}: The pertinent details are provided in Section~\ref{subsec-method}. The threshold is set to the 95-th percentile of all activation probabilities. For instance, in the case of LLaMA-2 (70B), threshold is established at 0.515. This stipulates that the neurons we select are required to exhibit an activation probability exceeding 0.515 for at least one language.

(b) \emph{Language activation probability (LAP)}: There are also methods to identify neurons directly. But most of them are infeasible due to the high computational cost~\cite{gurnee-etal-2024-universal,dai-etal-2022-knowledge}. Inspired by~\citet{voita-etal-2023-neurons}, we apply their method by identifying a neuron as language-specific if its activation probability exceeds 95\%.

(c) \emph{Language activation value entropy (LAVE)}: This is a variant of our proposed method, wherein we substitute the activation probability with the mean activation value across languages. Similarly, we normalize the activation value and calculate the entropy to find neurons with high activation value  in response to particular languages.

(d) \emph{Parameter variation (PV)}: By extending the work of \citet{zhao-etal-2023-unveiling}, we compare the model parameters before and after monolingual instruction tuning to identify language-specific parameters. 
In particular, we train individual models on the Alpaca instruction datasets~\cite{Taori-etal-2023-stanford} and its multilingual version~\cite{chen-etal-2023-multilingualsift} which comprise 52,000 instances for each target. These models undergo training for two epochs, with a batch size of 128 and a constant learning rate of \mbox{1e-5}. 
We mainly consider the parameters inside the MHA and FFN modules, \ie the weight matrices in Equations~\ref{eq-1},~\ref{eq-2}, and~\ref{eq-3}.
We compute the rate of change across various languages, and select parameters that exhibit a low rate of change in one or two languages but a high rate in others.
In detail, we refine the change ratio by subtracting the maximum value and then conduct L1 normalization for entropy calculation.

(e) \emph{Random selection (RS)}: Additionally, we add a baseline to randomly select neurons for each language, serving as a reference for different methods as shown in Figure~\ref{fig:random} in Appendix. 



\subsection{Main Perturbation Experiments}\label{subsec-results}

\begin{figure}[t!]
      \centering
	   \begin{subfigure}{0.49\linewidth}
		\includegraphics[width=\linewidth]{images/results0.pdf}
		\caption*{(a) LLaMA-2 (7B)}
	   \end{subfigure}
        \begin{subfigure}{0.49\linewidth}
		\includegraphics[width=\linewidth]{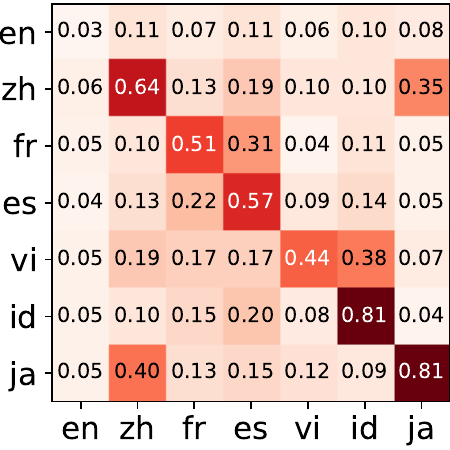}
		\caption*{(b) LLaMA-2-Chat (7B)}
	   \end{subfigure}
	   \begin{subfigure}{0.49\linewidth}
		\includegraphics[width=\linewidth]{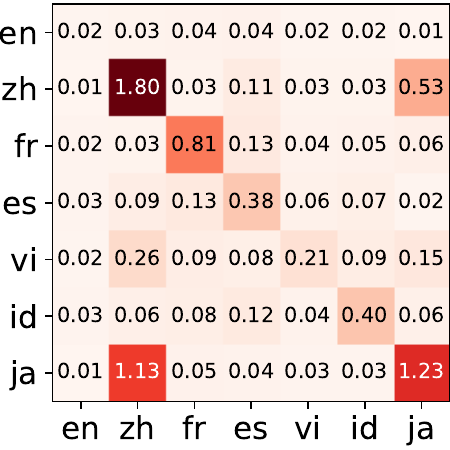}
		\caption*{(c) LLaMA-2 (13B)}
	    \end{subfigure}
	     \begin{subfigure}{0.49\linewidth}
		  \includegraphics[width=\linewidth]{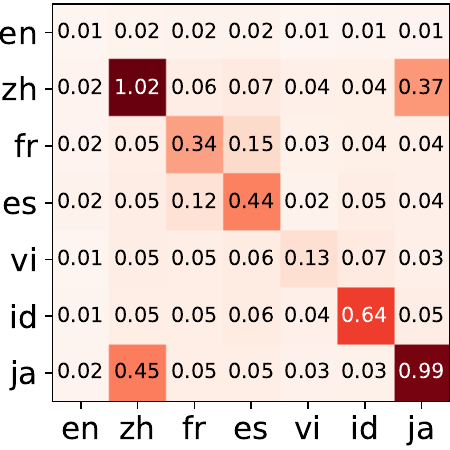}
		 \caption*{(d) LLaMA-2 (70B)}
	      \end{subfigure}
       
	       \begin{subfigure}{0.49\linewidth}
		  \includegraphics[width=\linewidth]{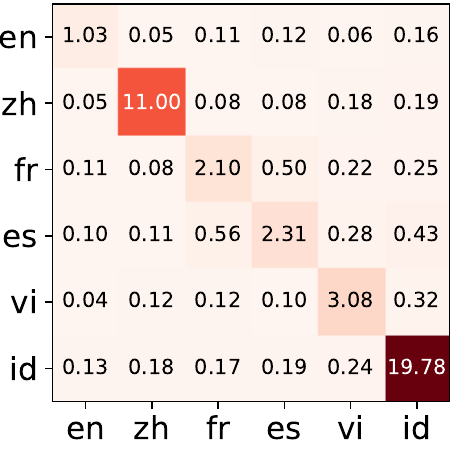}
		  \caption*{(e) BLOOM (7B)}
	       \end{subfigure}
         \begin{subfigure}{0.49\linewidth}
		  \includegraphics[width=\linewidth]{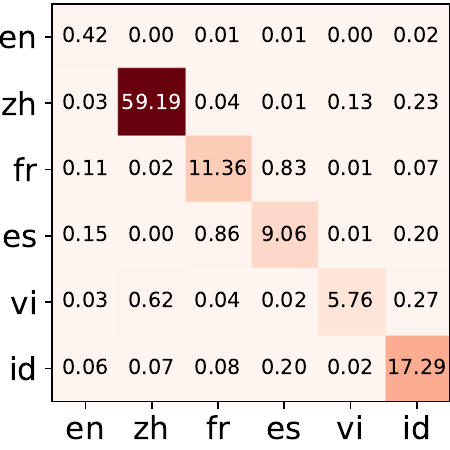}
		  \caption*{(f) OPT (6.7B)}
	       \end{subfigure}
         \begin{subfigure}{0.49\linewidth}
		  \includegraphics[width=\linewidth]{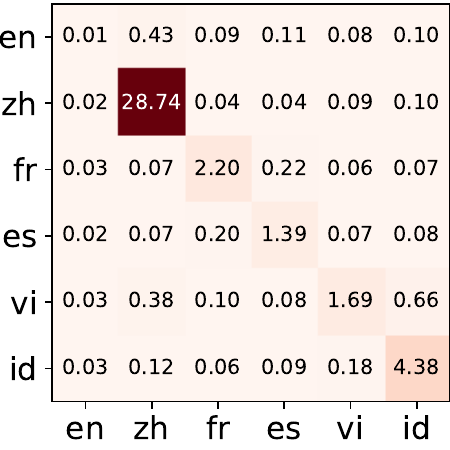}
		  \caption*{(g) Mistral (7B)}
	       \end{subfigure}
         \begin{subfigure}{0.49\linewidth}
		  \includegraphics[width=\linewidth]{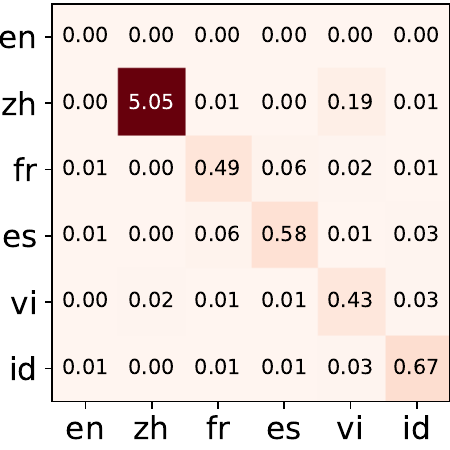}
		  \caption*{(h) Phi-2 (2.7B)}
	       \end{subfigure}
	\caption{Applying our LAPE method to different model types and sizes.}
	\label{fig:ppl-results2}
\end{figure}

In this part, we conduct the perturbation experiments by deactivating the identified language-specific regions. 
Specially, for all the comparison methods in Section~\ref{sec:exp-setup}, we identify 1\% of the neurons or parameters and treat them as language-specific regions. We then  set the activation values of these neurons to zero, or directly zero the parameters to assess the impact on the model's  multilingual capacities based on language modeling and open-ended text generation tasks.

Figure~\ref{fig:ppl-results1} presents the perturbation results (measured by PPL change on language modeling task) of different identification methods on LLaMA-2. Overall, we can see that large impact values for LAPE mainly occur in diagonal entries. 
It indicates that our LAPE method is adept at identifying language-specific neurons. Deactivating neurons associated with a specific language predominantly affects the PPL results of that language, with negligible effects on others. 
In contrast, the variant utilizing activation values (LAVE) causes cross-lingual interference (\eg the entry $\langle id, es \rangle$), and the rest methods fail to exhibit clear language-specific patterns. 

\begin{table}[t]
\centering
\resizebox{1.0\columnwidth}{!}{
\begin{tabular}{ccccccc}
\toprule
   & \textbf{zh}   & \textbf{fr}   & \textbf{es}   & \textbf{vi}   & \textbf{id}   & \textbf{ja}   \\ 
\midrule
\textbf{Normal} & 4.30 & 4.19 & 3.51 & 3.70 & 4.16 & 2.86 \\
\textbf{Random}     & 4.18 & 4.22 & 3.35 & 3.53 & 4.42 & 2.99 \\
\midrule
\textbf{zh} & \textbf{2.46} & 3.56 & 2.96 & 3.64 & 3.56 & 2.31 \\
\textbf{fr} & 3.69 & \textbf{2.50} & 2.29 & 3.01 & 3.59 & 2.76 \\
\textbf{es} & 3.51 & 2.57 & \textbf{2.01} & 3.14 & 3.34 & 2.56 \\
\textbf{vi} & 3.93 & 3.19 & 2.49 & \textbf{2.74} & 3.59 & 2.74 \\
\textbf{id} & 3.67 & 3.10 & 2.67 & 3.21 & \textbf{2.84} & 2.80 \\
\textbf{ja} & 3.21 & 3.69 & 3.07 & 3.49 & 3.37 & \textbf{1.84} \\ 
\bottomrule
\end{tabular}}
\caption{Performance of LLaMA-2 (70B) on the multilingual Vicuna as evaluated by GPT-4. The ``normal'' row is baseline scores without deactivation while the ``random'' row is with random deactivation. Subsequent rows are scores with deactivation of specific neurons.
}
\label{tab:mvicuna}
\end{table}

We further investigate whether our method is effective across different model sizes and model types. As depicted in Figure~\ref{fig:ppl-results2}, the language-specific degradation patterns are evident for LLaMA-2, BLOOM, OPT, Mistral, and Phi-2. 
An interesting find is that neurons in larger models tend to be specialized for a single language rather than being shared among two or more languages. 
Furthermore, we can find that there exists a high correlation between Chinese and Japanese: when we deactivating the neurons specific to one language, the performance of the other language would be affected (\eg the entries $\langle zh, ja \rangle$ and $\langle ja, zh \rangle$).  
By inspecting into the identified neurons of the two languages, we note that a substantial amount of neurons (approximately 25\%) actually overlap for both languages. It is likely because Chinese and Japanese partially share common characters.   
In addition, it can be observed that our LAPE method leads to similar findings on BLOOM and other LLMs, which further verify the generality of our proposed LAPE method.   

When employing LAPE to open-ended generation tasks based on the Vicuna dataset, we can clearly observe from Table~\ref{tab:mvicuna}  that deactivating language-specific neurons significantly impairs the generation capabilities in the target language. 
We further provide an illustrative example in Table~\ref{tab:example} of the model's response to a question in Simplified Chinese when the neurons associated with Simplified Chinese are deactivated.
We can see a chaotic mixture of Traditional Chinese characters and redundant English phrases, indicating a severe degradation of language capacity in Simplified Chinese. 




\begin{table}[!t]
\small
\centering
\begin{tabular}{p{0.95\columnwidth}}
    \toprule
    \textbf{Question} \\
    \begin{CJK*}{UTF8}{gbsn} 你是一位登上珠穆朗玛峰顶峰的登山者。描述一下你的情绪和从高处看到的景色。\end{CJK*} \\
    \textcolor{gray}{(\textit{Translation}: You are a mountain climber reaching the summit of Mount Everest. Describe your emotions and the view.)} \\
    \midrule
    \textbf{Normal output} \\
    \begin{CJK*}{UTF8}{gbsn}我是一个登上珠穆朗玛峰顶峰的登山者。当我站在山顶时，我感到非常兴奋和自豪。…\end{CJK*} \\
    \textcolor{gray}{(\textit{Translation}: I am a climber who has reached the summit of Mount Everest. When I stood on the top of the mountain, I felt very excited and proud. ...)} \\
    \midrule
    \textbf{Deactivated output} \\
    \begin{CJK*}{UTF8}{bsmi}我是一個登上珠穆朗瑪峰頂峰的登山者。 I am a mountaineer who has climbed to the top of Mount Everest. 當我站在珠my朗ma峰頂峰，我感到非常興奮和欣慰。 … \end{CJK*}\\
    \bottomrule                                            
\end{tabular}
\caption{Illustration of LLaMA-2-70B responses to a question in Simplified Chinese. The text in black is model's actual output and \textcolor{gray}{text in gray} is our self-added translation. The deactivated output is the generation when neurons of Simplified Chinese are deactivated.}
\label{tab:example}
\end{table}

%% file: sec-analysis.tex
\subsection{Further Analysis}
After presenting the main experiments, we further conduct detailed analysis experiments to investigate language-specific neurons and their impact. Unless specified, all analysis results are obtained based on LLaMA-2 (70B). 

\subsubsection{Distribution and Identification Ratio}




\begin{table}[ht]
  \centering
  \resizebox{1.0\columnwidth}{!}{
    \begin{tabular}{ccccccc}
    \toprule
    \textbf{en} & \textbf{zh} & \textbf{fr} & \textbf{es} & \textbf{vi} & \textbf{id} & \textbf{ja} \\
    \midrule
    836 & 5,153 & 6,082 & 6,154 & 4,980 & 6,106 & 5,216 \\
    \bottomrule
    \end{tabular}}
  \caption{The number of neurons in each language.}
  \label{tab:language}
\end{table}

\paragraph{Neuron distribution across languages.}
After running our LAPE method on LLaMA-2 (70B), we identify around 23K language-specific neurons. The distribution of these neurons across individual languages is detailed in Table~\ref{tab:language}. Since neurons may be shared by multiple languages, the sum of language-specific neurons actually surpass 23K. Overall, except English, the distribution of neurons is relatively even across languages. 
However, the number of English-specific neurons is much smaller than the other languages. We speculate that English is the dominant language in LLaMA-2, and thus it requires fewer neurons to support the specific language ability. 

\begin{figure}[t]
    \centering
    \includegraphics[width=0.95\columnwidth]{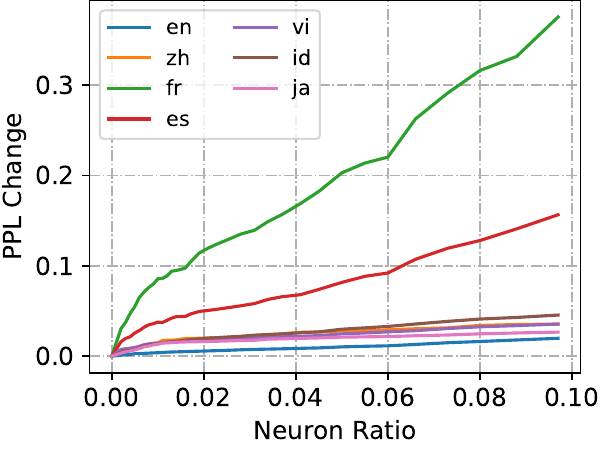}
    \caption{Change in PPL across different languages upon incremental number of language specific neurons when deactivating French neurons.}
    \label{fig-neuron-number}
\end{figure}

\paragraph{Increasing threshold ratios for identification.} 
In Section~\ref{subsec-results}, we consider a mere 1\% of the neurons as being language specific. We further vary the selection ratio of language-specific neurons from 0 to 10\%, to examine its impact on multilingual processing. 
Here, we select \emph{French} for study, while the results on the other languages are similar. 
When deactivating neurons specific to French, we observe a significant  increase in the PPL on French in Figure~\ref{fig-neuron-number}, while the impact on the rest languages are relatively limited with the exception of Spanish. 
it is consistent with our intuition: the performance of the being perturbed language and its related (or similar) language would be severely harmed. 


\subsubsection{Structural Distribution Analysis}  \label{subsec:at_bottom_top_layers}

In this part, we analyze how language-specific neurons are distributed across different layers.  

\begin{figure}[t]
    \centering
    \includegraphics[width=0.95\columnwidth]{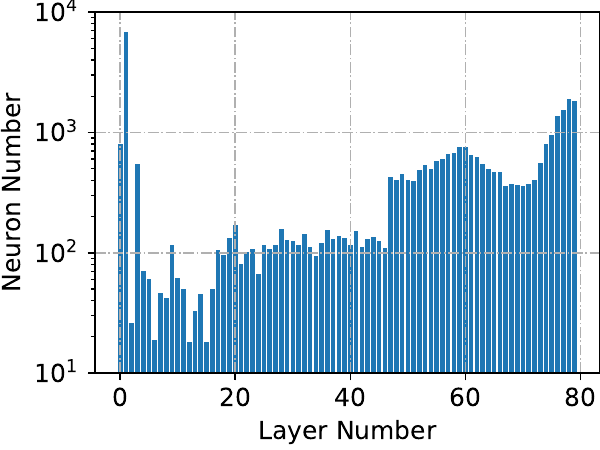}
    \caption{Distribution of language-specific neurons across different layers in LLaMA-2 (70B).}
    \label{fig-data-amount}
\end{figure}

\paragraph{Language processing is concentrated at bottom and top layers.}
In Figure~\ref{fig-data-amount}, we present the layer-wise distribution of language-specific neurons across various layers, which has a skewed ``U'' shape.  
This finding indicates that  language-specific neurons have a pronounced concentration in the top and bottom layers. 
Specifically, the second layer has approximately 7,000 language-specific neurons, while layers 5 through 47 only contain about 100 neurons each. Further, the neuron count gradually increases, with the final four layers each comprising over 1,000 neurons. The complete statistics about the layer-wise distribution across various languages are reported in Table~\ref{tab:activation_prob_llama70b}. 

\begin{figure}[t]
\centering
\includegraphics[width=0.95\columnwidth]{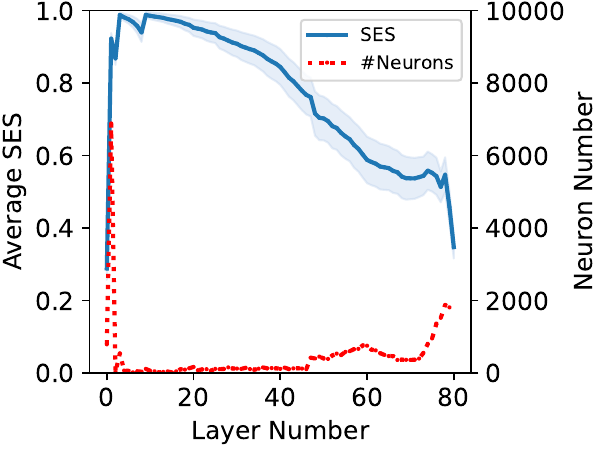}
\caption{The mean SES between all language pairs and total language neuron numbers across layers.}
\label{fig:at_bottom_top}
\end{figure}

\paragraph{Why such a skewed  distribution?} 
To understand why such a skewed distribution occurs, we seek explanations from multilingual semantic representation by exploring how multilingual aligned texts are represented across the layers. 
Specially, we employ the multilingual Vicuna dataset (Section~\ref{sec:exp-setup}), comprising of aligned texts in different languages. Given a group of aligned texts, we feed them into the LLM and obtain the sentence embedding of each text for each layer. 
We then compute the \emph{mean sentence embedding similarity (SES)} between each pair of the aligned texts across languages in Figure~\ref{fig:at_bottom_top}.
Interestingly, the SES curve shows an opposite trend with the distribution of language-specific neurons. At the beginning, the similarity quickly increases, then reaches the peak, and gradually decreases to a small value. 
This finding suggests that: at the bottom layers, the LLM needs to map aligned texts of different languages into the shared representation space, thus requiring more language-specific neurons for semantic transformation; while at top layers serving for token generation, the LLM needs to handle vocabulary mapping, which also requires the support of more language-specific neurons. 

\subsubsection{Language Dominance Analysis}\label{subsec:cl}

Since the high-resource language (\ie English) in the LLaMA-2 training corpus has a surprisingly smaller number of neurons than other languages, as indicated in Table~\ref{tab:language}, we speculate that there might exist some dominance relations between high-resource and low-resource languages, which depends on the composition of pre-training corpus. 


\paragraph{Language dominance measurement.}
Inspired by \citet{xu-etal-2023-language-representation}, we transfer the sentence embeddings across different languages into the same space around a target language, and  examine how texts from the other languages are aligned to the texts of the target language. 
Firstly, for each language $k$, we compute the mean sentence embeddings of all its texts, and obtain $\bm{v}_k^i$ as the language vector of $k$ at $i$-th layer. 
Then we follow the same formula proposed by \citet{xu-etal-2023-language-representation} to conduct the space mapping for each text from language $k$:
\begin{equation}\label{eq:hki}
    \hat{\bm{h}}_k^i = \bm{h}_k^i-\bm{v}_k^i+\bm{v}_{c}^i,
\end{equation}
where $\hat{\bm{h}}_k^i$ denotes a transformed  embedding of some text in the $i$-th layer of language $k$. Here, we specify $c$ as the target language, and compute the mean sentence embedding similarity (SES, Section~\ref{subsec:at_bottom_top_layers}) over all the sentence pairs between languages $k$ and $c$, based on the transformed representations in Eq.~\ref{eq:hki}. A larger SES score indicates language $c$ has a larger dominance degree. 

\begin{figure}[t]
    \begin{subfigure}{0.49\linewidth}
        \includegraphics[width=\linewidth]{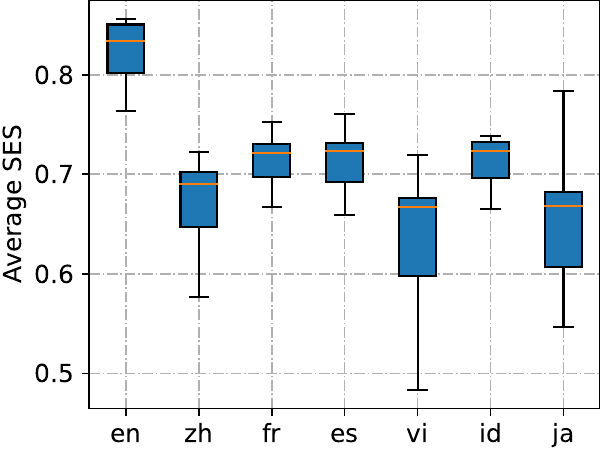}
        \caption{LLaMA-2 (70B)}
        \label{fig:cl-llama}
   \end{subfigure}
   \begin{subfigure}{0.49\linewidth}
        \includegraphics[width=\linewidth]{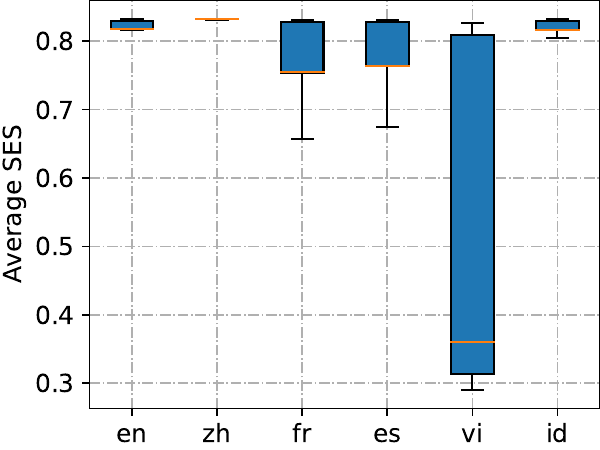}
        \caption{BLOOM (170B)}
        \label{fig:cl-bloom}
   \end{subfigure}
\centering
\caption{Dominance scores (mean SES)  across layers when different languages serve as the target language.}
\label{fig:cl}
\end{figure}

\paragraph{Low-resource languages are centered around high-resource languages.}
To compute the dominance degree, we still use the  multilingual Vicuna dataset (Section~\ref{sec:exp-setup}). 
From the results of LLaMA-2 (70B) in Figure~\ref{fig:cl-llama}, we can see that the mean SES is obviously higher than all other languages when the target language is English. 
As English is the high-resource language of LLaMA-2,  low-resource languages need to be aligned with English for achieving  better performance. 
When it comes to BLOOM (170B) in Figure~\ref{fig:cl-bloom}, several languages (\eg English and Chinese) show dominance, since it is originally pre-trained on multilingual corpora. 













\begin{table}[t]
\centering
\resizebox{1.0\columnwidth}{!}{
\begin{tabular}{cccccccc}
\toprule
\textbf{Metrics} & \textbf{Settings}  &  \textbf{zh}   & \textbf{fr}   & \textbf{es}   & \textbf{vi}   & \textbf{id}   & \textbf{ja}   \\ 
\midrule
Language & normal  & 0.87 & 0.73 & 0.81 & 0.60 & 0.40 & 0.79 \\
\cline{2-8}
accuracy & steered & 0.99 & 0.90 & 0.93 & 0.97 & 0.99 & 1.00 \\
\midrule
Content & normal  & 4.30 & 4.19 & 3.51 & 3.70 & 4.16 & 2.86 \\
\cline{2-8}
quality & steered & 4.57 & 4.35 & 4.02 & 3.57 & 4.28 & 2.91 \\
\bottomrule
\end{tabular}}
\caption{The language accuracy and content score of the normal output and the steered output by activating language-specific neurons. Language accuracy is computed by whether the model responds in a given language using the \texttt{langdetect} package and the content quality is measured by GPT-4.}
\label{tab:steering}
\end{table}

\subsubsection{Case Study}
Finally, we explore the possibility to ``steer'' the output language of LLMs to mitigate the \emph{off-target} problem and facilitate cross-lingual generation.

Researchers have observed that when prompting in one language, language models may generate responses in a different language, such a phenomenon referred to as the \emph{off-target language issue}~\cite{gu-etal-2019-improved,sennrich-etal-2023-mitigating}. We speculate that language-specific neuron might not be activated in this case. Thus, we attempt to manually activate these neurons by increasing their activation value to the average activation value of this language. As shown in Table~\ref{tab:steering}, it is evident that the model demonstrates a significantly higher propensity to respond in the expected language, as indicated by the language accuracy. Additionally, the generation quality, as assessed by GPT-4 metrics, shows remarkable improvement.

Moreover, we attempt to steer cross-lingual generation by posing a question in one language and answering it in another. As presented in Table~\ref{tab:case2}, we deactivate Spanish neurons and activate Chinese neurons when replying to a question in Spanish. The model successfully generates fluent Chinese text, showing the feasibility of altering the output language through neuron-level manipulation.

\begin{table}[t!]
\small
\centering
\begin{tabular}{p{0.95\columnwidth}}
    \toprule
    \textbf{Question} \\
    ?`C\'{o}mo puedo mejorar mis habilidades de gesti\'{o}n del tiempo? \\
    \textcolor{gray}{(\textit{Translation}: How can I improve my time management skills?)} \\
    \midrule
    \textbf{Steered output in Chinese} \\
    \begin{CJK*}{UTF8}{gbsn}Time管理是一个非常重要的技能，因为它可以提高你的生产力并使你更有效地完成任务。下面是一些提示来改进你的时间管理技能：1、制定计划：每天开始时制定清单并将任务分配给特定时间段。这样可以确保你在每天都有足够的时间来完成所有任务。2、设置目标：设置明确而可实现的目标可以帮助你保持专注并提高效果。3、抵制分心：当你正在工作时，请不要被其他事情分心。4、利用技术：……\end{CJK*}\\
    \textcolor{gray}{(\textit{Translation}: Time management is a very important skill because it increases your productivity and allows you to complete tasks more efficiently. Here are some tips to improve your time management skills: 1. Make a plan: Make a list at the beginning of each day and assign tasks to specific time periods. This ensures that you have enough time each day to complete all your tasks. 2. Set goals: Setting clear and achievable goals can help you stay focused and improve your results. 3. Resist distractions: When you are working, please don’t be distracted by other things. 4. Utilize technology: ...)} \\
    \bottomrule                                            
\end{tabular}
\caption{An illustrative example of asking question in Spanish and answering it by deactivating Spanish neurons and activating Chinese neurons.}
\label{tab:case2}
\end{table}

%% file: sec-rel.tex
\section{Related Work}
\paragraph{Multilingual ability.}
Multilingual ability has been widely studied since the emergence of pre-trained language models~\cite{devlin-etal-2019-bert,xue-etal-2021-mt5}. Recently, LLMs~\cite{zhao-etal-2023-survey,nguyen-etal-2023-seallms} showcase more excellent multilingual capabilities even without explicit language alignment~\cite{kulshreshtha-etal-2020-cross,cao-etal-2020-multilingual}. 
Extensive research has been made to investigate the factors that influence models' multilingual ability~\cite{philippy-etal-2023-towards}. For example, linguistic similarity has been examined~\cite{pires-etal-2019-multilingual,dufter-schutze-2020-identifying,wang-etal-2019-cross,conneau-etal-2020-emerging}, which is generally believed to correlate with models' cross-lingual ability. Specially, ``word order'' shows some contradictions about whether it really affects multilingual ability~\cite{pires-etal-2019-multilingual,deshpande-etal-2022-bert}. Not only limited to language property, training settings have also been considered~\cite{lauscher-etal-2020-zero,ahuja-etal-2022-multi}. 

Existing work has explored language-agnostic (or language-shared) components within multilingual models.
For example, researchers concentrate on shared knowledge across various languages~\cite{stanczak-etal-2022-neurons,chen-etal-2023-journey,zhao-etal-2023-unveiling,bricken-etal-2023-monosemanticity}. 
However, the exploration of language-specific components within LLMs remains an under-investigated area.





\paragraph{Neuron analysis.}
Neuron analysis has gained significant attention in recent years, paralleling research in neurobiological studies of the human brain~\cite{friederici-etal-2011-brain,parr-etal-2022-active}. Originating from vision models~\cite{bau-etal-2020-understanding,mu-etal-2020-compositional}, neuron analysis views neuron activation as the recall of learned knowledge~\cite{sajjad-etal-2022-neuron}. Researchers widely adopt these methods to analyze the sources of specific abilities or skills in language models, including sentiment analysis~\citep{radford-etal-2017-learning}, knowledge storage~\citep{dai-etal-2022-knowledge}, and task-solving~\cite{wang-etal-2022-finding-skill}.

Recent studies have also discovered that certain neurons can convey specialized contexts~\cite{gurnee-etal-2023-finding,bills-etal-2023-language}, such as positional information~\cite{voita-etal-2023-neurons} and linguistic properties~\cite{bau-etal-2018-identifying,xin-etal-2019-part,dalvi-etal-2019-one,dalvi-etal-2020-analyzing}. Moreover, \citet{gurnee-etal-2024-universal} utilize Pearson correlation to calculate neuron similarity, identifying some universal neurons across models. In contrast to previous research, we have developed a method applicable to LLMs that unveils the mechanism of their multilingual abilities. This approach offers a more practical and effective solution for neuron analysis in multilingual scenarios.

%% file: sec-app.tex
\clearpage
\appendix

\section{Appendix}
Table~\ref{tab:languages} compiles the statistics of pre-training corpora in LLaMA-2~\cite{touvron-etal-2023-llama2} and BLOOM~\cite{workshop-etal-2022-bloom}.

We list the number of language-specific neurons across different layers of BLOOM (7B), LLaMA-2 (7B), LLaMA-2 (13B), and LLaMA-2 (70B) in Tables~\ref{tab:activation_prob_bloom}, \ref{tab:activation_prob_llama7b}, \ref{tab:activation_prob_llama13b}, and \ref{tab:activation_prob_llama70b}.

\begin{table}[htbp]
  \centering
  \resizebox{0.9\columnwidth}{!}{
    \begin{tabular}{ccccccc}
    \toprule
   \#Layer  & en & zh & fr & es & vi & id\\
   \midrule
1 & 0 & 0 & 0 & 0 & 0 & 0\\
2 & 0 & 0 & 0 & 0 & 0 & 0\\
3 & 0 & 0 & 0 & 0 & 0 & 0\\
4 & 0 & 0 & 0 & 0 & 0 & 0\\
5 & 0 & 1 & 0 & 0 & 0 & 0\\
6 & 0 & 0 & 0 & 0 & 0 & 0\\
7 & 0 & 1 & 0 & 0 & 1 & 0\\
8 & 0 & 0 & 0 & 0 & 1 & 0\\
9 & 0 & 0 & 1 & 0 & 0 & 0\\
10 & 1 & 2 & 1 & 1 & 0 & 0\\
11 & 0 & 0 & 2 & 2 & 1 & 1\\
12 & 0 & 0 & 0 & 0 & 0 & 0\\
13 & 1 & 1 & 1 & 1 & 0 & 0\\
14 & 0 & 1 & 1 & 1 & 0 & 0\\
15 & 1 & 2 & 1 & 0 & 0 & 0\\
16 & 1 & 0 & 2 & 2 & 1 & 2\\
17 & 3 & 0 & 1 & 1 & 0 & 0\\
18 & 2 & 1 & 3 & 2 & 0 & 1\\
19 & 3 & 4 & 3 & 4 & 2 & 2\\
20 & 1 & 1 & 2 & 1 & 1 & 1\\
21 & 11 & 7 & 7 & 8 & 3 & 4\\
22 & 8 & 8 & 9 & 9 & 7 & 9\\
23 & 9 & 19 & 11 & 12 & 12 & 15\\
24 & 21 & 24 & 24 & 24 & 26 & 46\\
25 & 24 & 34 & 24 & 28 & 35 & 90\\
26 & 24 & 37 & 47 & 54 & 40 & 180\\
27 & 34 & 46 & 66 & 93 & 70 & 330\\
28 & 62 & 79 & 106 & 151 & 83 & 562\\
29 & 86 & 126 & 155 & 240 & 103 & 817\\
30 & 153 & 259 & 213 & 284 & 165 & 763\\
    \bottomrule
    \end{tabular}%
    }
  \caption{Neuron number per layer of BLOOM (7B).}
  \label{tab:activation_prob_bloom}%
\end{table}%

\begin{table}[htbp]
  \centering
  \resizebox{\columnwidth}{!}{
    \begin{tabular}{ccccc}
    \toprule
    Language & Code & Family & BLOOM Ratio & LLaMA-2 Ratio \\
    \midrule
    English & en & Indo-European & 33.68\% & 89.70\% \\
    Chinese & zh & Sino-Tibetan & 18.13\% & 0.13\% \\
    French & fr & Indo-European & 14.46\% & 0.16\% \\
    Spanish & es & Indo-European & 12.16\% & 0.13\% \\
    Vietnamese & vi & Austro-Asiatic & 3.04\% & 0.08\% \\
    Indonesian & id & Austronesian & 1.39\% & 0.03\% \\
    Japanese & ja & Japonic & 0.00\% & 0.10\% \\
    \bottomrule
    \end{tabular}%
    }
  \caption{The language statistics of the pre-training corpora in BLOOM and LLaMA-2.}
  \label{tab:languages}%
\end{table}

\begin{table}[htbp]
  \centering
  \resizebox{0.9\columnwidth}{!}{
    \begin{tabular}{cccccccc}
    \toprule
    \#Layer & en & zh & fr & es & vi & id & ja\\
    \midrule
    1 & 17 & 108 & 220 & 195 & 274 & 221 & 109\\
    2 & 0 & 32 & 39 & 27 & 16 & 15 & 31\\
    3 & 0 & 1 & 2 & 2 & 0 & 2 & 0\\
    4 & 0 & 0 & 0 & 0 & 0 & 0 & 0\\
    5 & 2 & 3 & 6 & 4 & 4 & 4 & 3\\
    6 & 3 & 5 & 5 & 4 & 2 & 3 & 4\\
    7 & 0 & 9 & 10 & 8 & 8 & 4 & 4\\
    8 & 1 & 5 & 1 & 1 & 3 & 1 & 3\\
    9 & 0 & 2 & 1 & 0 & 1 & 1 & 3\\
    10 & 0 & 3 & 3 & 4 & 3 & 4 & 5\\
    11 & 0 & 5 & 1 & 0 & 5 & 2 & 6\\
    12 & 3 & 7 & 5 & 4 & 3 & 0 & 6\\
    13 & 1 & 8 & 10 & 8 & 11 & 8 & 11\\
    14 & 2 & 19 & 7 & 5 & 16 & 8 & 18\\
    15 & 1 & 13 & 12 & 10 & 13 & 9 & 15\\
    16 & 1 & 7 & 3 & 1 & 5 & 4 & 15\\
    17 & 3 & 28 & 17 & 14 & 15 & 12 & 20\\
    18 & 3 & 11 & 13 & 11 & 19 & 16 & 18\\
    19 & 1 & 17 & 6 & 7 & 16 & 13 & 21\\
    20 & 2 & 20 & 18 & 8 & 20 & 24 & 26\\
    21 & 3 & 19 & 33 & 15 & 35 & 29 & 32\\
    22 & 3 & 22 & 21 & 23 & 26 & 49 & 13\\
    23 & 0 & 33 & 60 & 42 & 38 & 84 & 35\\
    24 & 2 & 20 & 56 & 31 & 49 & 84 & 18\\
    25 & 0 & 20 & 78 & 58 & 33 & 77 & 19\\
    26 & 3 & 11 & 80 & 54 & 30 & 78 & 17\\
    27 & 2 & 18 & 86 & 72 & 43 & 88 & 7\\
    28 & 2 & 14 & 50 & 59 & 35 & 64 & 13\\
    29 & 5 & 15 & 49 & 48 & 36 & 58 & 14\\
    30 & 7 & 23 & 44 & 39 & 27 & 40 & 17\\
    31 & 18 & 36 & 54 & 52 & 31 & 38 & 29\\
    32 & 10 & 49 & 32 & 32 & 19 & 28 & 50\\
    
    \bottomrule
    \end{tabular}%
    }
  \caption{Neuron number per layer of LLaMA-2 (7B).}
  \label{tab:activation_prob_llama7b}%
\end{table}

\begin{table}[htbp]
  \centering
  \resizebox{0.9\columnwidth}{!}{
    \begin{tabular}{cccccccc}
    \toprule
\#Layer & en & zh & fr & es & vi & id & ja\\
\midrule
1 & 60 & 127 & 222 & 189 & 248 & 184 & 206\\
2 & 9 & 162 & 177 & 118 & 187 & 69 & 305\\
3 & 0 & 2 & 1 & 2 & 1 & 2 & 1\\
4 & 0 & 1 & 1 & 1 & 0 & 2 & 2\\
5 & 0 & 3 & 0 & 1 & 0 & 0 & 3\\
6 & 1 & 3 & 2 & 2 & 3 & 3 & 5\\
7 & 0 & 5 & 1 & 1 & 4 & 3 & 6\\
8 & 3 & 7 & 3 & 3 & 3 & 1 & 9\\
9 & 2 & 18 & 7 & 7 & 10 & 3 & 9\\
10 & 0 & 12 & 9 & 6 & 8 & 5 & 9\\
11 & 0 & 15 & 18 & 17 & 11 & 8 & 12\\
12 & 2 & 5 & 3 & 2 & 5 & 3 & 9\\
13 & 1 & 7 & 2 & 1 & 3 & 2 & 9\\
14 & 0 & 5 & 3 & 2 & 4 & 0 & 10\\
15 & 1 & 7 & 3 & 3 & 7 & 5 & 8\\
16 & 3 & 25 & 20 & 14 & 22 & 11 & 31\\
17 & 3 & 30 & 16 & 11 & 28 & 21 & 32\\
18 & 4 & 40 & 40 & 31 & 35 & 25 & 47\\
19 & 1 & 26 & 23 & 13 & 22 & 19 & 44\\
20 & 1 & 24 & 14 & 16 & 14 & 9 & 35\\
21 & 1 & 28 & 19 & 17 & 22 & 17 & 34\\
22 & 3 & 43 & 26 & 13 & 40 & 37 & 55\\
23 & 3 & 32 & 23 & 10 & 22 & 24 & 49\\
24 & 1 & 28 & 20 & 12 & 20 & 32 & 27\\
25 & 1 & 24 & 29 & 8 & 23 & 25 & 29\\
26 & 3 & 27 & 40 & 32 & 27 & 42 & 31\\
27 & 2 & 40 & 63 & 41 & 31 & 64 & 36\\
28 & 4 & 20 & 50 & 43 & 21 & 48 & 30\\
29 & 2 & 25 & 78 & 48 & 19 & 71 & 26\\
30 & 0 & 25 & 89 & 88 & 38 & 75 & 19\\
31 & 2 & 21 & 72 & 52 & 46 & 77 & 13\\
32 & 3 & 16 & 83 & 60 & 36 & 80 & 14\\
33 & 0 & 23 & 69 & 55 & 31 & 61 & 19\\
34 & 1 & 27 & 47 & 54 & 35 & 69 & 16\\
35 & 1 & 20 & 69 & 58 & 41 & 67 & 23\\
36 & 1 & 21 & 60 & 54 & 42 & 58 & 27\\
37 & 5 & 22 & 33 & 32 & 31 & 47 & 11\\
38 & 14 & 40 & 58 & 52 & 48 & 44 & 36\\
39 & 8 & 57 & 43 & 35 & 26 & 27 & 51\\
40 & 15 & 105 & 47 & 51 & 38 & 44 & 97\\

    \bottomrule
    \end{tabular}%
    }
  \caption{Neuron number per layer of LLaMA-2 (13B).}
  \label{tab:activation_prob_llama13b}%
\end{table}%

\begin{table}[htbp]
  \centering 
  \resizebox{0.75\columnwidth}{!}{
    \begin{tabular}{cccccccc}
    \toprule
\#Layer & en & zh & fr & es & vi & id & ja\\
\midrule
1 & 238 & 199 & 45 & 43 & 28 & 47 & 195\\
2 & 117 & 886 & 1056 & 1155 & 1589 & 897 & 1184\\
3 & 2 & 5 & 2 & 3 & 5 & 5 & 4\\
4 & 11 & 79 & 105 & 69 & 62 & 108 & 109\\
5 & 0 & 17 & 9 & 6 & 12 & 12 & 15\\
6 & 5 & 10 & 11 & 10 & 10 & 7 & 8\\
7 & 2 & 3 & 2 & 3 & 3 & 2 & 4\\
8 & 1 & 14 & 5 & 5 & 6 & 4 & 11\\
9 & 2 & 9 & 7 & 6 & 5 & 5 & 8\\
10 & 2 & 25 & 23 & 15 & 17 & 8 & 25\\
11 & 0 & 13 & 14 & 10 & 8 & 6 & 11\\
12 & 0 & 16 & 5 & 6 & 7 & 5 & 11\\
13 & 0 & 5 & 2 & 1 & 2 & 2 & 6\\
14 & 0 & 9 & 3 & 2 & 7 & 2 & 10\\
15 & 0 & 14 & 3 & 3 & 8 & 4 & 13\\
16 & 0 & 4 & 1 & 4 & 3 & 1 & 5\\
17 & 1 & 13 & 7 & 8 & 7 & 5 & 9\\
18 & 3 & 22 & 12 & 13 & 16 & 10 & 29\\
19 & 2 & 28 & 13 & 11 & 11 & 8 & 22\\
20 & 4 & 34 & 19 & 21 & 18 & 11 & 26\\
21 & 1 & 38 & 27 & 19 & 26 & 25 & 33\\
22 & 1 & 16 & 17 & 15 & 10 & 7 & 14\\
23 & 1 & 23 & 17 & 14 & 15 & 14 & 18\\
24 & 1 & 18 & 15 & 14 & 26 & 13 & 20\\
25 & 2 & 10 & 11 & 11 & 9 & 11 & 12\\
26 & 1 & 23 & 12 & 15 & 17 & 14 & 35\\
27 & 4 & 28 & 13 & 11 & 10 & 12 & 29\\
28 & 3 & 25 & 14 & 16 & 20 & 17 & 20\\
29 & 6 & 39 & 23 & 21 & 19 & 19 & 30\\
30 & 0 & 24 & 23 & 23 & 19 & 19 & 20\\
31 & 1 & 15 & 30 & 24 & 20 & 22 & 13\\
32 & 2 & 21 & 17 & 20 & 23 & 16 & 18\\
33 & 2 & 21 & 22 & 17 & 23 & 27 & 32\\
34 & 1 & 20 & 19 & 13 & 18 & 23 & 17\\
35 & 0 & 14 & 12 & 17 & 18 & 19 & 14\\
36 & 3 & 17 & 22 & 16 & 20 & 23 & 19\\
37 & 4 & 26 & 29 & 18 & 24 & 24 & 30\\
38 & 4 & 17 & 31 & 24 & 16 & 19 & 20\\
39 & 2 & 18 & 27 & 26 & 17 & 23 & 26\\
40 & 4 & 20 & 26 & 15 & 23 & 21 & 24\\
41 & 2 & 17 & 15 & 14 & 20 & 25 & 24\\
42 & 2 & 21 & 26 & 22 & 22 & 30 & 28\\
43 & 0 & 21 & 13 & 15 & 17 & 17 & 28\\
44 & 1 & 17 & 18 & 14 & 23 & 31 & 25\\
45 & 1 & 24 & 23 & 11 & 22 & 30 & 23\\
46 & 1 & 17 & 21 & 11 & 18 & 33 & 24\\
47 & 2 & 13 & 22 & 14 & 14 & 27 & 18\\
48 & 1 & 55 & 78 & 64 & 55 & 93 & 77\\
49 & 2 & 54 & 68 & 61 & 55 & 73 & 90\\
50 & 4 & 61 & 85 & 100 & 42 & 98 & 65\\
51 & 0 & 49 & 97 & 66 & 43 & 80 & 69\\
52 & 3 & 49 & 99 & 80 & 37 & 83 & 40\\
53 & 1 & 64 & 120 & 96 & 55 & 86 & 64\\
54 & 0 & 55 & 136 & 130 & 54 & 116 & 46\\
55 & 1 & 55 & 118 & 109 & 52 & 114 & 49\\
56 & 4 & 62 & 134 & 130 & 74 & 135 & 44\\
57 & 0 & 47 & 149 & 162 & 64 & 132 & 50\\
58 & 1 & 40 & 187 & 172 & 73 & 142 & 45\\
59 & 0 & 38 & 162 & 184 & 87 & 155 & 43\\
60 & 4 & 59 & 190 & 208 & 84 & 163 & 54\\
61 & 1 & 57 & 178 & 180 & 80 & 211 & 47\\
62 & 2 & 39 & 142 & 165 & 75 & 180 & 43\\
63 & 2 & 45 & 137 & 160 & 72 & 174 & 36\\
64 & 2 & 36 & 123 & 138 & 76 & 138 & 37\\
65 & 0 & 40 & 104 & 123 & 58 & 145 & 31\\
66 & 3 & 35 & 90 & 112 & 67 & 124 & 39\\
67 & 4 & 51 & 86 & 103 & 69 & 112 & 43\\
68 & 1 & 27 & 63 & 74 & 55 & 101 & 40\\
69 & 3 & 33 & 64 & 69 & 73 & 85 & 44\\
70 & 6 & 39 & 67 & 66 & 56 & 81 & 51\\
71 & 8 & 55 & 60 & 58 & 65 & 69 & 47\\
72 & 4 & 50 & 75 & 75 & 55 & 64 & 47\\
73 & 10 & 74 & 62 & 60 & 53 & 87 & 55\\
74 & 18 & 94 & 84 & 82 & 88 & 107 & 84\\
75 & 15 & 154 & 132 & 154 & 98 & 140 & 113\\
76 & 30 & 188 & 139 & 152 & 122 & 178 & 148\\
77 & 37 & 254 & 244 & 239 & 162 & 242 & 186\\
78 & 57 & 292 & 245 & 255 & 177 & 270 & 230\\
79 & 89 & 450 & 256 & 263 & 192 & 226 & 402\\
80 & 81 & 484 & 219 & 220 & 179 & 192 & 438\\

    \bottomrule
    \end{tabular}%
    }
  \caption{Neuron number per layer of LLaMA-2 (70B).}
  \label{tab:activation_prob_llama70b}%
\end{table}%

\begin{figure}[t]
    \centering
    \includegraphics[width=0.5\columnwidth]{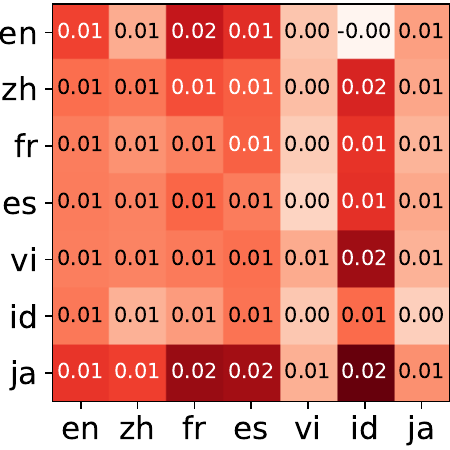}
    \caption{The results of deactivating neurons randomly.}
    \label{fig:random}
\end{figure}

%% file: main.bbl
\begin{thebibliography}{55}
\expandafter\ifx\csname natexlab\endcsname\relax\def\natexlab#1{#1}\fi

\bibitem[{Achiam et~al.(2023)Achiam, Adler, Agarwal, Ahmad, Akkaya, Aleman, Almeida, Altenschmidt, Altman, Anadkat et~al.}]{achiam-etal-2023-gpt}
Josh Achiam, Steven Adler, Sandhini Agarwal, Lama Ahmad, Ilge Akkaya, Florencia~Leoni Aleman, Diogo Almeida, Janko Altenschmidt, Sam Altman, Shyamal Anadkat, et~al. 2023.
\newblock Gpt-4 technical report.
\newblock \emph{arXiv preprint arXiv:2303.08774}.

\bibitem[{Ahuja et~al.(2022)Ahuja, Kumar, Dandapat, and Choudhury}]{ahuja-etal-2022-multi}
Kabir Ahuja, Shanu Kumar, Sandipan Dandapat, and Monojit Choudhury. 2022.
\newblock \href {https://doi.org/10.18653/v1/2022.acl-long.374} {Multi task learning for zero shot performance prediction of multilingual models}.
\newblock In \emph{Proceedings of the 60th Annual Meeting of the Association for Computational Linguistics (Volume 1: Long Papers)}, pages 5454--5467, Dublin, Ireland. Association for Computational Linguistics.

\bibitem[{Anil et~al.(2023)Anil, Dai, Firat, Johnson, Lepikhin, Passos, Shakeri, Taropa, Bailey, Chen et~al.}]{anil-etal-2023-palm}
Rohan Anil, Andrew~M Dai, Orhan Firat, Melvin Johnson, Dmitry Lepikhin, Alexandre Passos, Siamak Shakeri, Emanuel Taropa, Paige Bailey, Zhifeng Chen, et~al. 2023.
\newblock Palm 2 technical report.
\newblock \emph{arXiv preprint arXiv:2305.10403}.

\bibitem[{Bau et~al.(2018)Bau, Belinkov, Sajjad, Durrani, Dalvi, and Glass}]{bau-etal-2018-identifying}
Anthony Bau, Yonatan Belinkov, Hassan Sajjad, Nadir Durrani, Fahim Dalvi, and James Glass. 2018.
\newblock Identifying and controlling important neurons in neural machine translation.
\newblock \emph{arXiv preprint arXiv:1811.01157}.

\bibitem[{Bau et~al.(2020)Bau, Zhu, Strobelt, Lapedriza, Zhou, and Torralba}]{bau-etal-2020-understanding}
David Bau, Jun-Yan Zhu, Hendrik Strobelt, Agata Lapedriza, Bolei Zhou, and Antonio Torralba. 2020.
\newblock Understanding the role of individual units in a deep neural network.
\newblock \emph{Proceedings of the National Academy of Sciences}, 117(48):30071--30078.

\bibitem[{Bills et~al.(2023)Bills, Cammarata, Mossing, Tillman, Gao, Goh, Sutskever, Leike, Wu, and Saunders}]{bills-etal-2023-language}
Steven Bills, Nick Cammarata, Dan Mossing, Henk Tillman, Leo Gao, Gabriel Goh, Ilya Sutskever, Jan Leike, Jeff Wu, and William Saunders. 2023.
\newblock Language models can explain neurons in language models.
\newblock \emph{URL https://openaipublic. blob. core. windows. net/neuron-explainer/paper/index. html.(Date accessed: 14.05. 2023)}.

\bibitem[{Bricken et~al.(2023)Bricken, Templeton, Batson, Chen, Jermyn, Conerly, Turner, Anil, Denison, Askell, Lasenby, Wu, Kravec, Schiefer, Maxwell, Joseph, Hatfield-Dodds, Tamkin, Nguyen, McLean, Burke, Hume, Carter, Henighan, and Olah}]{bricken-etal-2023-monosemanticity}
Trenton Bricken, Adly Templeton, Joshua Batson, Brian Chen, Adam Jermyn, Tom Conerly, Nick Turner, Cem Anil, Carson Denison, Amanda Askell, Robert Lasenby, Yifan Wu, Shauna Kravec, Nicholas Schiefer, Tim Maxwell, Nicholas Joseph, Zac Hatfield-Dodds, Alex Tamkin, Karina Nguyen, Brayden McLean, Josiah~E Burke, Tristan Hume, Shan Carter, Tom Henighan, and Christopher Olah. 2023.
\newblock Towards monosemanticity: Decomposing language models with dictionary learning.
\newblock \emph{Transformer Circuits Thread}.
\newblock Https://transformer-circuits.pub/2023/monosemantic-features/index.html.

\bibitem[{Cao et~al.(2020)Cao, Kitaev, and Klein}]{cao-etal-2020-multilingual}
Steven Cao, Nikita Kitaev, and Dan Klein. 2020.
\newblock Multilingual alignment of contextual word representations.
\newblock \emph{arXiv preprint arXiv:2002.03518}.

\bibitem[{Chen et~al.(2023{\natexlab{a}})Chen, Cao, Chen, Liu, and Zhao}]{chen-etal-2023-journey}
Yuheng Chen, Pengfei Cao, Yubo Chen, Kang Liu, and Jun Zhao. 2023{\natexlab{a}}.
\newblock Journey to the center of the knowledge neurons: Discoveries of language-independent knowledge neurons and degenerate knowledge neurons.
\newblock \emph{arXiv preprint arXiv:2308.13198}.

\bibitem[{Chen et~al.(2023{\natexlab{b}})Chen, Yan, Liang, Jiang, Wu, Yu, Chen, Chen, Zhang, Jianquan, Xiang, and Wang}]{chen-etal-2023-multilingualsift}
Zhihong Chen, Shuo Yan, Juhao Liang, Feng Jiang, Xiangbo Wu, Fei Yu, Guiming~Hardy Chen, Junying Chen, Hongbo Zhang, Li~Jianquan, Wan Xiang, and Benyou Wang. 2023{\natexlab{b}}.
\newblock \href {https://github.com/FreedomIntelligence/MultilingualSIFT.git} {{MultilingualSIFT: Multilingual Supervised Instruction Fine-tuning}}.

\bibitem[{Chiang et~al.(2023)Chiang, Li, Lin, Sheng, Wu, Zhang, Zheng, Zhuang, Zhuang, Gonzalez, Stoica, and Xing}]{weilin-etal-2023-vicuna}
Wei-Lin Chiang, Zhuohan Li, Zi~Lin, Ying Sheng, Zhanghao Wu, Hao Zhang, Lianmin Zheng, Siyuan Zhuang, Yonghao Zhuang, Joseph~E. Gonzalez, Ion Stoica, and Eric~P. Xing. 2023.
\newblock \href {https://lmsys.org/blog/2023-03-30-vicuna/} {Vicuna: An open-source chatbot impressing gpt-4 with 90\%* chatgpt quality}.

\bibitem[{Conneau et~al.(2020{\natexlab{a}})Conneau, Khandelwal, Goyal, Chaudhary, Wenzek, Guzm{\'a}n, Grave, Ott, Zettlemoyer, and Stoyanov}]{conneau-etal-2020-unsupervised}
Alexis Conneau, Kartikay Khandelwal, Naman Goyal, Vishrav Chaudhary, Guillaume Wenzek, Francisco Guzm{\'a}n, Edouard Grave, Myle Ott, Luke Zettlemoyer, and Veselin Stoyanov. 2020{\natexlab{a}}.
\newblock \href {https://doi.org/10.18653/v1/2020.acl-main.747} {Unsupervised cross-lingual representation learning at scale}.
\newblock In \emph{Proceedings of the 58th Annual Meeting of the Association for Computational Linguistics}, pages 8440--8451, Online. Association for Computational Linguistics.

\bibitem[{Conneau et~al.(2020{\natexlab{b}})Conneau, Wu, Li, Zettlemoyer, and Stoyanov}]{conneau-etal-2020-emerging}
Alexis Conneau, Shijie Wu, Haoran Li, Luke Zettlemoyer, and Veselin Stoyanov. 2020{\natexlab{b}}.
\newblock \href {https://doi.org/10.18653/v1/2020.acl-main.536} {Emerging cross-lingual structure in pretrained language models}.
\newblock In \emph{Proceedings of the 58th Annual Meeting of the Association for Computational Linguistics}, pages 6022--6034, Online. Association for Computational Linguistics.

\bibitem[{Dai et~al.(2022)Dai, Dong, Hao, Sui, Chang, and Wei}]{dai-etal-2022-knowledge}
Damai Dai, Li~Dong, Yaru Hao, Zhifang Sui, Baobao Chang, and Furu Wei. 2022.
\newblock \href {https://doi.org/10.18653/v1/2022.acl-long.581} {Knowledge neurons in pretrained transformers}.
\newblock In \emph{Proceedings of the 60th Annual Meeting of the Association for Computational Linguistics (Volume 1: Long Papers)}, pages 8493--8502, Dublin, Ireland. Association for Computational Linguistics.

\bibitem[{Dalvi et~al.(2019)Dalvi, Durrani, Sajjad, Belinkov, Bau, and Glass}]{dalvi-etal-2019-one}
Fahim Dalvi, Nadir Durrani, Hassan Sajjad, Yonatan Belinkov, Anthony Bau, and James Glass. 2019.
\newblock What is one grain of sand in the desert? analyzing individual neurons in deep nlp models.
\newblock In \emph{Proceedings of the AAAI Conference on Artificial Intelligence}, volume~33, pages 6309--6317.

\bibitem[{Dalvi et~al.(2020)Dalvi, Sajjad, Durrani, and Belinkov}]{dalvi-etal-2020-analyzing}
Fahim Dalvi, Hassan Sajjad, Nadir Durrani, and Yonatan Belinkov. 2020.
\newblock Analyzing redundancy in pretrained transformer models.
\newblock \emph{arXiv preprint arXiv:2004.04010}.

\bibitem[{Deshpande et~al.(2022)Deshpande, Talukdar, and Narasimhan}]{deshpande-etal-2022-bert}
Ameet Deshpande, Partha Talukdar, and Karthik Narasimhan. 2022.
\newblock \href {https://doi.org/10.18653/v1/2022.naacl-main.264} {When is {BERT} multilingual? isolating crucial ingredients for cross-lingual transfer}.
\newblock In \emph{Proceedings of the 2022 Conference of the North American Chapter of the Association for Computational Linguistics: Human Language Technologies}, pages 3610--3623, Seattle, United States. Association for Computational Linguistics.

\bibitem[{Devlin et~al.(2019)Devlin, Chang, Lee, and Toutanova}]{devlin-etal-2019-bert}
Jacob Devlin, Ming-Wei Chang, Kenton Lee, and Kristina Toutanova. 2019.
\newblock \href {https://doi.org/10.18653/v1/N19-1423} {{BERT}: Pre-training of deep bidirectional transformers for language understanding}.
\newblock In \emph{Proceedings of the 2019 Conference of the North {A}merican Chapter of the Association for Computational Linguistics: Human Language Technologies, Volume 1 (Long and Short Papers)}, pages 4171--4186, Minneapolis, Minnesota. Association for Computational Linguistics.

\bibitem[{Dufter and Sch{\"u}tze(2020)}]{dufter-schutze-2020-identifying}
Philipp Dufter and Hinrich Sch{\"u}tze. 2020.
\newblock \href {https://doi.org/10.18653/v1/2020.emnlp-main.358} {Identifying elements essential for {BERT}{'}s multilinguality}.
\newblock In \emph{Proceedings of the 2020 Conference on Empirical Methods in Natural Language Processing (EMNLP)}, pages 4423--4437, Online. Association for Computational Linguistics.

\bibitem[{Friederici(2011)}]{friederici-etal-2011-brain}
Angela~D Friederici. 2011.
\newblock The brain basis of language processing: from structure to function.
\newblock \emph{Physiological reviews}, 91(4):1357--1392.

\bibitem[{Gu et~al.(2019)Gu, Wang, Cho, and Li}]{gu-etal-2019-improved}
Jiatao Gu, Yong Wang, Kyunghyun Cho, and Victor~O.K. Li. 2019.
\newblock \href {https://doi.org/10.18653/v1/P19-1121} {Improved zero-shot neural machine translation via ignoring spurious correlations}.
\newblock In \emph{Proceedings of the 57th Annual Meeting of the Association for Computational Linguistics}, pages 1258--1268, Florence, Italy. Association for Computational Linguistics.

\bibitem[{Gurnee et~al.(2024)Gurnee, Horsley, Guo, Kheirkhah, Sun, Hathaway, Nanda, and Bertsimas}]{gurnee-etal-2024-universal}
Wes Gurnee, Theo Horsley, Zifan~Carl Guo, Tara~Rezaei Kheirkhah, Qinyi Sun, Will Hathaway, Neel Nanda, and Dimitris Bertsimas. 2024.
\newblock Universal neurons in gpt2 language models.
\newblock \emph{arXiv preprint arXiv:2401.12181}.

\bibitem[{Gurnee et~al.(2023)Gurnee, Nanda, Pauly, Harvey, Troitskii, and Bertsimas}]{gurnee-etal-2023-finding}
Wes Gurnee, Neel Nanda, Matthew Pauly, Katherine Harvey, Dmitrii Troitskii, and Dimitris Bertsimas. 2023.
\newblock Finding neurons in a haystack: Case studies with sparse probing.
\newblock \emph{arXiv preprint arXiv:2305.01610}.

\bibitem[{Hendrycks and Gimpel(2016)}]{hendrycks-etal-2016-gaussian}
Dan Hendrycks and Kevin Gimpel. 2016.
\newblock Gaussian error linear units (gelus).
\newblock \emph{arXiv preprint arXiv:1606.08415}.

\bibitem[{Javaheripi et~al.(2023)Javaheripi, Bubeck, Abdin, Aneja, Bubeck, Mendes, Chen, Del~Giorno, Eldan, Gopi et~al.}]{javaheripi-etal-2023-phi}
Mojan Javaheripi, S{\'e}bastien Bubeck, Marah Abdin, Jyoti Aneja, Sebastien Bubeck, Caio C{\'e}sar~Teodoro Mendes, Weizhu Chen, Allie Del~Giorno, Ronen Eldan, Sivakanth Gopi, et~al. 2023.
\newblock Phi-2: The surprising power of small language models.
\newblock \emph{Microsoft Research Blog}.

\bibitem[{Jiang et~al.(2023)Jiang, Sablayrolles, Mensch, Bamford, Chaplot, Casas, Bressand, Lengyel, Lample, Saulnier et~al.}]{jiang-etal-2023-mistral}
Albert~Q Jiang, Alexandre Sablayrolles, Arthur Mensch, Chris Bamford, Devendra~Singh Chaplot, Diego de~las Casas, Florian Bressand, Gianna Lengyel, Guillaume Lample, Lucile Saulnier, et~al. 2023.
\newblock Mistral 7b.
\newblock \emph{arXiv preprint arXiv:2310.06825}.

\bibitem[{Khanna et~al.(2024)Khanna, Mu{\~n}oz, Kim, Kfir, Paulk, Jamali, Cai, Mustroph, Caprara, Hardstone et~al.}]{khanna-etal-2024-single}
Arjun~R Khanna, William Mu{\~n}oz, Young~Joon Kim, Yoav Kfir, Angelique~C Paulk, Mohsen Jamali, Jing Cai, Martina~L Mustroph, Irene Caprara, Richard Hardstone, et~al. 2024.
\newblock Single-neuronal elements of speech production in humans.
\newblock \emph{Nature}, pages 1--8.

\bibitem[{Kulshreshtha et~al.(2020)Kulshreshtha, Redondo-Garc{\'\i}a, and Chang}]{kulshreshtha-etal-2020-cross}
Saurabh Kulshreshtha, Jos{\'e}~Luis Redondo-Garc{\'\i}a, and Ching-Yun Chang. 2020.
\newblock Cross-lingual alignment methods for multilingual bert: A comparative study.
\newblock \emph{arXiv preprint arXiv:2009.14304}.

\bibitem[{Lauscher et~al.(2020)Lauscher, Ravishankar, Vuli{\'c}, and Glava{\v{s}}}]{lauscher-etal-2020-zero}
Anne Lauscher, Vinit Ravishankar, Ivan Vuli{\'c}, and Goran Glava{\v{s}}. 2020.
\newblock \href {https://doi.org/10.18653/v1/2020.emnlp-main.363} {From zero to hero: {O}n the limitations of zero-shot language transfer with multilingual {T}ransformers}.
\newblock In \emph{Proceedings of the 2020 Conference on Empirical Methods in Natural Language Processing (EMNLP)}, pages 4483--4499, Online. Association for Computational Linguistics.

\bibitem[{Mu and Andreas(2020)}]{mu-etal-2020-compositional}
Jesse Mu and Jacob Andreas. 2020.
\newblock Compositional explanations of neurons.
\newblock \emph{Advances in Neural Information Processing Systems}, 33:17153--17163.

\bibitem[{Nair and Hinton(2010)}]{Nair-etal-2010-rectified}
Vinod Nair and Geoffrey~E. Hinton. 2010.
\newblock Rectified linear units improve restricted boltzmann machines.
\newblock In \emph{Proceedings of the 27th International Conference on International Conference on Machine Learning}, ICML'10, page 807–814, Madison, WI, USA. Omnipress.

\bibitem[{Nguyen et~al.(2023)Nguyen, Zhang, Li, Aljunied, Tan, Cheng, Chen, Deng, Yang, Liu et~al.}]{nguyen-etal-2023-seallms}
Xuan-Phi Nguyen, Wenxuan Zhang, Xin Li, Mahani Aljunied, Qingyu Tan, Liying Cheng, Guanzheng Chen, Yue Deng, Sen Yang, Chaoqun Liu, et~al. 2023.
\newblock Seallms--large language models for southeast asia.
\newblock \emph{arXiv preprint arXiv:2312.00738}.

\bibitem[{Parr et~al.(2022)Parr, Pezzulo, and Friston}]{parr-etal-2022-active}
Thomas Parr, Giovanni Pezzulo, and Karl~J Friston. 2022.
\newblock \emph{Active inference: the free energy principle in mind, brain, and behavior}.
\newblock MIT Press.

\bibitem[{Philippy et~al.(2023)Philippy, Guo, and Haddadan}]{philippy-etal-2023-towards}
Fred Philippy, Siwen Guo, and Shohreh Haddadan. 2023.
\newblock \href {https://doi.org/10.18653/v1/2023.acl-long.323} {Towards a common understanding of contributing factors for cross-lingual transfer in multilingual language models: A review}.
\newblock In \emph{Proceedings of the 61st Annual Meeting of the Association for Computational Linguistics (Volume 1: Long Papers)}, pages 5877--5891, Toronto, Canada. Association for Computational Linguistics.

\bibitem[{Pires et~al.(2019)Pires, Schlinger, and Garrette}]{pires-etal-2019-multilingual}
Telmo Pires, Eva Schlinger, and Dan Garrette. 2019.
\newblock \href {https://doi.org/10.18653/v1/P19-1493} {How multilingual is multilingual {BERT}?}
\newblock In \emph{Proceedings of the 57th Annual Meeting of the Association for Computational Linguistics}, pages 4996--5001, Florence, Italy. Association for Computational Linguistics.

\bibitem[{Radford et~al.(2017)Radford, Jozefowicz, and Sutskever}]{radford-etal-2017-learning}
Alec Radford, Rafal Jozefowicz, and Ilya Sutskever. 2017.
\newblock Learning to generate reviews and discovering sentiment.
\newblock \emph{arXiv preprint arXiv:1704.01444}.

\bibitem[{Sajjad et~al.(2022)Sajjad, Durrani, and Dalvi}]{sajjad-etal-2022-neuron}
Hassan Sajjad, Nadir Durrani, and Fahim Dalvi. 2022.
\newblock \href {https://doi.org/10.1162/tacl_a_00519} {Neuron-level interpretation of deep {NLP} models: A survey}.
\newblock \emph{Transactions of the Association for Computational Linguistics}, 10:1285--1303.

\bibitem[{Scao et~al.(2022)Scao, Fan, Akiki, Pavlick, Ili{\'c}, Hesslow, Castagn{\'e}, Luccioni, Yvon et~al.}]{workshop-etal-2022-bloom}
Teven~Le Scao, Angela Fan, Christopher Akiki, Ellie Pavlick, Suzana Ili{\'c}, Daniel Hesslow, Roman Castagn{\'e}, Alexandra~Sasha Luccioni, Fran{\c{c}}ois Yvon, et~al. 2022.
\newblock Bloom: A 176b-parameter open-access multilingual language model.
\newblock \emph{arXiv preprint arXiv:2211.05100}.

\bibitem[{Sennrich et~al.(2023)Sennrich, Vamvas, and Mohammadshahi}]{sennrich-etal-2023-mitigating}
Rico Sennrich, Jannis Vamvas, and Alireza Mohammadshahi. 2023.
\newblock Mitigating hallucinations and off-target machine translation with source-contrastive and language-contrastive decoding.
\newblock \emph{arXiv preprint arXiv:2309.07098}.

\bibitem[{Shazeer(2020)}]{shazeer-etal-2020-glu}
Noam Shazeer. 2020.
\newblock Glu variants improve transformer.
\newblock \emph{arXiv preprint arXiv:2002.05202}.

\bibitem[{Stanczak et~al.(2022)Stanczak, Ponti, Torroba~Hennigen, Cotterell, and Augenstein}]{stanczak-etal-2022-neurons}
Karolina Stanczak, Edoardo Ponti, Lucas Torroba~Hennigen, Ryan Cotterell, and Isabelle Augenstein. 2022.
\newblock \href {https://doi.org/10.18653/v1/2022.naacl-main.114} {Same neurons, different languages: Probing morphosyntax in multilingual pre-trained models}.
\newblock In \emph{Proceedings of the 2022 Conference of the North American Chapter of the Association for Computational Linguistics: Human Language Technologies}, pages 1589--1598, Seattle, United States. Association for Computational Linguistics.

\bibitem[{Taori et~al.(2023)Taori, Gulrajani, Zhang, Dubois, Li, Guestrin, Liang, and Hashimoto}]{Taori-etal-2023-stanford}
Rohan Taori, Ishaan Gulrajani, Tianyi Zhang, Yann Dubois, Xuechen Li, Carlos Guestrin, Percy Liang, and Tatsunori~B. Hashimoto. 2023.
\newblock Stanford alpaca: An instruction-following llama model.
\newblock \url{https://github.com/tatsu-lab/stanford_alpaca}.

\bibitem[{Touvron et~al.(2023{\natexlab{a}})Touvron, Lavril, Izacard, Martinet, Lachaux, Lacroix, Rozi{\`e}re, Goyal, Hambro, Azhar et~al.}]{touvron-etal-2023-llama}
Hugo Touvron, Thibaut Lavril, Gautier Izacard, Xavier Martinet, Marie-Anne Lachaux, Timoth{\'e}e Lacroix, Baptiste Rozi{\`e}re, Naman Goyal, Eric Hambro, Faisal Azhar, et~al. 2023{\natexlab{a}}.
\newblock Llama: Open and efficient foundation language models.
\newblock \emph{arXiv preprint arXiv:2302.13971}.

\bibitem[{Touvron et~al.(2023{\natexlab{b}})Touvron, Martin, Stone, Albert, Almahairi, Babaei, Bashlykov, Batra, Bhargava, Bhosale et~al.}]{touvron-etal-2023-llama2}
Hugo Touvron, Louis Martin, Kevin Stone, Peter Albert, Amjad Almahairi, Yasmine Babaei, Nikolay Bashlykov, Soumya Batra, Prajjwal Bhargava, Shruti Bhosale, et~al. 2023{\natexlab{b}}.
\newblock Llama 2: Open foundation and fine-tuned chat models.
\newblock \emph{arXiv preprint arXiv:2307.09288}.

\bibitem[{Vaswani et~al.(2017)Vaswani, Shazeer, Parmar, Uszkoreit, Jones, Gomez, Kaiser, and Polosukhin}]{Vaswani-etal-2017-attention}
Ashish Vaswani, Noam Shazeer, Niki Parmar, Jakob Uszkoreit, Llion Jones, Aidan~N Gomez, \L~ukasz Kaiser, and Illia Polosukhin. 2017.
\newblock \href {https://proceedings.neurips.cc/paper_files/paper/2017/file/3f5ee243547dee91fbd053c1c4a845aa-Paper.pdf} {Attention is all you need}.
\newblock In \emph{Advances in Neural Information Processing Systems}, volume~30. Curran Associates, Inc.

\bibitem[{Voita et~al.(2023)Voita, Ferrando, and Nalmpantis}]{voita-etal-2023-neurons}
Elena Voita, Javier Ferrando, and Christoforos Nalmpantis. 2023.
\newblock Neurons in large language models: Dead, n-gram, positional.
\newblock \emph{arXiv preprint arXiv:2309.04827}.

\bibitem[{Wang et~al.(2022)Wang, Wen, Zhang, Hou, Liu, and Li}]{wang-etal-2022-finding-skill}
Xiaozhi Wang, Kaiyue Wen, Zhengyan Zhang, Lei Hou, Zhiyuan Liu, and Juanzi Li. 2022.
\newblock \href {https://doi.org/10.18653/v1/2022.emnlp-main.765} {Finding skill neurons in pre-trained transformer-based language models}.
\newblock In \emph{Proceedings of the 2022 Conference on Empirical Methods in Natural Language Processing}, pages 11132--11152, Abu Dhabi, United Arab Emirates. Association for Computational Linguistics.

\bibitem[{Wang et~al.(2019)Wang, Mayhew, Roth et~al.}]{wang-etal-2019-cross}
Zihan Wang, Stephen Mayhew, Dan Roth, et~al. 2019.
\newblock Cross-lingual ability of multilingual bert: An empirical study.
\newblock \emph{arXiv preprint arXiv:1912.07840}.

\bibitem[{Xin et~al.(2019)Xin, Lin, and Yu}]{xin-etal-2019-part}
Ji~Xin, Jimmy Lin, and Yaoliang Yu. 2019.
\newblock What part of the neural network does this? understanding lstms by measuring and dissecting neurons.
\newblock In \emph{Proceedings of the 2019 Conference on Empirical Methods in Natural Language Processing and the 9th International Joint Conference on Natural Language Processing (EMNLP-IJCNLP)}, pages 5823--5830.

\bibitem[{Xu et~al.(2023)Xu, Li, and Xiong}]{xu-etal-2023-language-representation}
Shaoyang Xu, Junzhuo Li, and Deyi Xiong. 2023.
\newblock \href {https://doi.org/10.18653/v1/2023.emnlp-main.226} {Language representation projection: Can we transfer factual knowledge across languages in multilingual language models?}
\newblock In \emph{Proceedings of the 2023 Conference on Empirical Methods in Natural Language Processing}, pages 3692--3702, Singapore. Association for Computational Linguistics.

\bibitem[{Xue et~al.(2021)Xue, Constant, Roberts, Kale, Al-Rfou, Siddhant, Barua, and Raffel}]{xue-etal-2021-mt5}
Linting Xue, Noah Constant, Adam Roberts, Mihir Kale, Rami Al-Rfou, Aditya Siddhant, Aditya Barua, and Colin Raffel. 2021.
\newblock \href {https://doi.org/10.18653/v1/2021.naacl-main.41} {m{T}5: A massively multilingual pre-trained text-to-text transformer}.
\newblock In \emph{Proceedings of the 2021 Conference of the North American Chapter of the Association for Computational Linguistics: Human Language Technologies}, pages 483--498, Online. Association for Computational Linguistics.

\bibitem[{Zhang et~al.(2022)Zhang, Roller, Goyal, Artetxe, Chen, Chen, Dewan, Diab, Li, Lin et~al.}]{zhang-etal-2022-opt}
Susan Zhang, Stephen Roller, Naman Goyal, Mikel Artetxe, Moya Chen, Shuohui Chen, Christopher Dewan, Mona Diab, Xian Li, Xi~Victoria Lin, et~al. 2022.
\newblock Opt: Open pre-trained transformer language models.
\newblock \emph{arXiv preprint arXiv:2205.01068}.

\bibitem[{Zhao et~al.(2023{\natexlab{a}})Zhao, Zhang, Ma, Zhang, Gui, Gao, and Huang}]{zhao-etal-2023-unveiling}
Jun Zhao, Zhihao Zhang, Yide Ma, Qi~Zhang, Tao Gui, Luhui Gao, and Xuanjing Huang. 2023{\natexlab{a}}.
\newblock Unveiling a core linguistic region in large language models.
\newblock \emph{arXiv preprint arXiv:2310.14928}.

\bibitem[{Zhao et~al.(2023{\natexlab{b}})Zhao, Zhou, Li, Tang, Wang, Hou, Min, Zhang, Zhang, Dong et~al.}]{zhao-etal-2023-survey}
Wayne~Xin Zhao, Kun Zhou, Junyi Li, Tianyi Tang, Xiaolei Wang, Yupeng Hou, Yingqian Min, Beichen Zhang, Junjie Zhang, Zican Dong, et~al. 2023{\natexlab{b}}.
\newblock A survey of large language models.
\newblock \emph{arXiv preprint arXiv:2303.18223}.

\bibitem[{Zheng et~al.(2023)Zheng, Chiang, Sheng, Zhuang, Wu, Zhuang, Lin, Li, Li, Xing et~al.}]{zheng-etal-2023-judging}
Lianmin Zheng, Wei-Lin Chiang, Ying Sheng, Siyuan Zhuang, Zhanghao Wu, Yonghao Zhuang, Zi~Lin, Zhuohan Li, Dacheng Li, Eric Xing, et~al. 2023.
\newblock Judging llm-as-a-judge with mt-bench and chatbot arena.
\newblock \emph{arXiv preprint arXiv:2306.05685}.

\end{thebibliography}
